\documentclass[journal]{IEEEtran}
\usepackage{amsmath,amsfonts}
\usepackage{algorithmic}
\usepackage{algorithm}
\usepackage{array}
\usepackage[caption=false,font=normalsize,labelfont=sf,textfont=sf]{subfig}
\usepackage{textcomp}
\usepackage{stfloats}
\usepackage{url}
\usepackage{verbatim}
\usepackage{graphicx}
\usepackage{cite}
\usepackage[table]{xcolor}
\usepackage{booktabs}
\usepackage{multirow}

\begin{document}

\author{%
    Lijia Yu, Jiuxin Cao, Yuchen Qiang, Changhao Chen, Yifei Huang, and Bo Liu%
    \thanks{Lijia Yu, Yuchen Qiang, Changhao Chen, and Yifei Huang are with the School of Cyber Science and Engineering, Southeast University, China (e-mail: lj\_yu@seu.edu.cn; yc.qiang@seu.edu.cn; chen\_changhao@seu.edu.cn; huang\_yifei@seu.edu.cn).}%
    \thanks{Jiuxin Cao is with the School of Cyber Science and Engineering, Southeast University, China, and also with Purple Mountain Laboratories, Nanjing, China (e-mail: jx.cao@seu.edu.cn).}%
    \thanks{Bo Liu is with the School of Computer Science and Engineering, Southeast University, China, and also with Purple Mountain Laboratories, Nanjing, China (e-mail: bliu@seu.edu.cn).}%
}

\title{Improving Adversarial Transferability on Vision-Language Pre-training Models via Surrogate-Specific Bias Correction}

\markboth{Journal of \LaTeX\ Class Files,~Vol.~14, No.~8, August~2021}%
{Shell \MakeLowercase{\textit{et al.}}: A Sample Article Using IEEEtran.cls for IEEE Journals}

\IEEEpubid{0000--0000/00\$00.00~\copyright~2021 IEEE}

\maketitle

\begin{abstract}
Adversarial examples reveal vulnerabilities in Vision-Language Pre-training (VLP) models and provide insights for improving robustness. A key property of adversarial examples is cross-model transferability, which enables transfer-based black-box attacks.
However, existing transfer-based attacks often rely heavily on the surrogate model, causing cross-model performance drops. One key reason is that adversarial optimization may follow surrogate model responses more than input semantics, making the update direction effective on the surrogate but less transferable to unseen targets. We call this dependency surrogate-specific bias.
Motivated by this observation, DeBias-Attack improves cross-model transferability by correcting this bias in adversarial optimization directions.
It maintains two perturbation branches. The main branch optimizes a perturbation on the original image and obtains the main adversarial gradient used to disrupt image-text alignment. The reference branch optimizes a perturbation on a weak-semantic image constructed from the dataset mean image by adding small Gaussian noise that is resampled at each iteration. Because this weak-semantic image contains little clear visual content, its optimization reflects surrogate responses more than image semantics, and its reference adversarial gradient estimates surrogate-specific bias. DeBias-Attack removes the aligned projection of the main gradient on the reference gradient before updating the adversarial image. It then performs context-aware text substitution using the updated adversarial image, yielding adversarial image-text pairs that transfer better across models.
DeBias-Attack is the first transfer-based VLP attack that corrects surrogate-specific bias through gradient correction. Experiments show strong performance across VLP models, downstream tasks, and both open-source and closed-source multimodal large language models.
\end{abstract}

\begin{IEEEkeywords}
Adversarial transferability, VLP models, black-box attack, transfer-based attack.
\end{IEEEkeywords}

\begin{figure}[!t]
    \centering
    \setlength{\abovecaptionskip}{0pt}
    \includegraphics[width=\columnwidth]{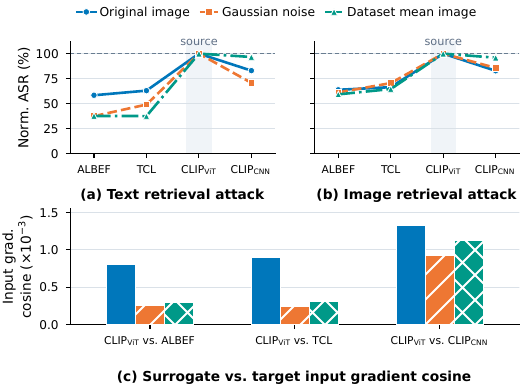}\par
    \vspace*{1pt}
    \caption{Observation with different CLIP$_{\rm ViT}$ optimization inputs: original image, Gaussian noise, and dataset mean image. (a)--(b) show normalized ASR for text retrieval and image retrieval attacks. (c) shows input gradient cosine between the CLIP$_{\rm ViT}$ surrogate and each target model.}
    \label{fig:observation}
\end{figure}

\section{Introduction}
\label{sec:introduction}
\IEEEPARstart{V}ision-Language Pre-training (VLP) models learn joint visual and textual representations from large-scale image-text pairs~\cite{li2021align}. They support image-text retrieval, image captioning, and visual grounding, and now form a common backbone for multimodal systems~\cite{li2022blip, li2023blip}. However, wider deployment has also exposed security risks~\cite{madry2017towards, lu2023set, goodfellow2014explaining, li2020bert, yang2022vision, gu2024a}. Among these risks, adversarial attacks on VLP models have drawn increasing attention, as maliciously crafted multimodal inputs can degrade predictions and undermine the reliability of downstream systems~\cite{zhang2022towards, luo2024image}.

\IEEEpubidadjcol

Among different attack settings, the black-box setting is especially important in practice, since attackers usually have no access to the internal architecture, parameters, or gradients of the target model~\cite{wang2021dual, zhang2022enhancing, zhao2023revisiting}. Transfer-based attacks address this setting by crafting adversarial examples on a surrogate model and applying them to unseen targets without victim interaction. This attack paradigm has been studied in unimodal and multimodal domains~\cite{wang2022generating, wang2021feature}. For VLP models, the strict surrogate-only setting is harder. Since the attacker receives no target feedback, attack success depends entirely on how well surrogate-crafted examples transfer.

However, improving adversarial transferability in this setting remains a fundamental challenge. Studies on deep models link transferability to the gradient directions used during adversarial optimization~\cite{zhao2021success, demontis2019adversarial}. From a first-order view, a surrogate-crafted perturbation transfers when its attack direction also increases the target-model loss, which depends on surrogate-target gradient alignment. If an update direction contains strong surrogate-specific components, it can attack the surrogate while becoming unstable on heterogeneous targets. In transfer-based VLP attacks, this risk is amplified because perturbations are optimized on the surrogate to disrupt image-text alignment, where coupled visual-textual representations can make the update direction follow surrogate-model responses. We study whether such surrogate-specific bias limits VLP transferability.

\begin{figure*}[!t]
    \centering
    \includegraphics[width=0.90\textwidth]{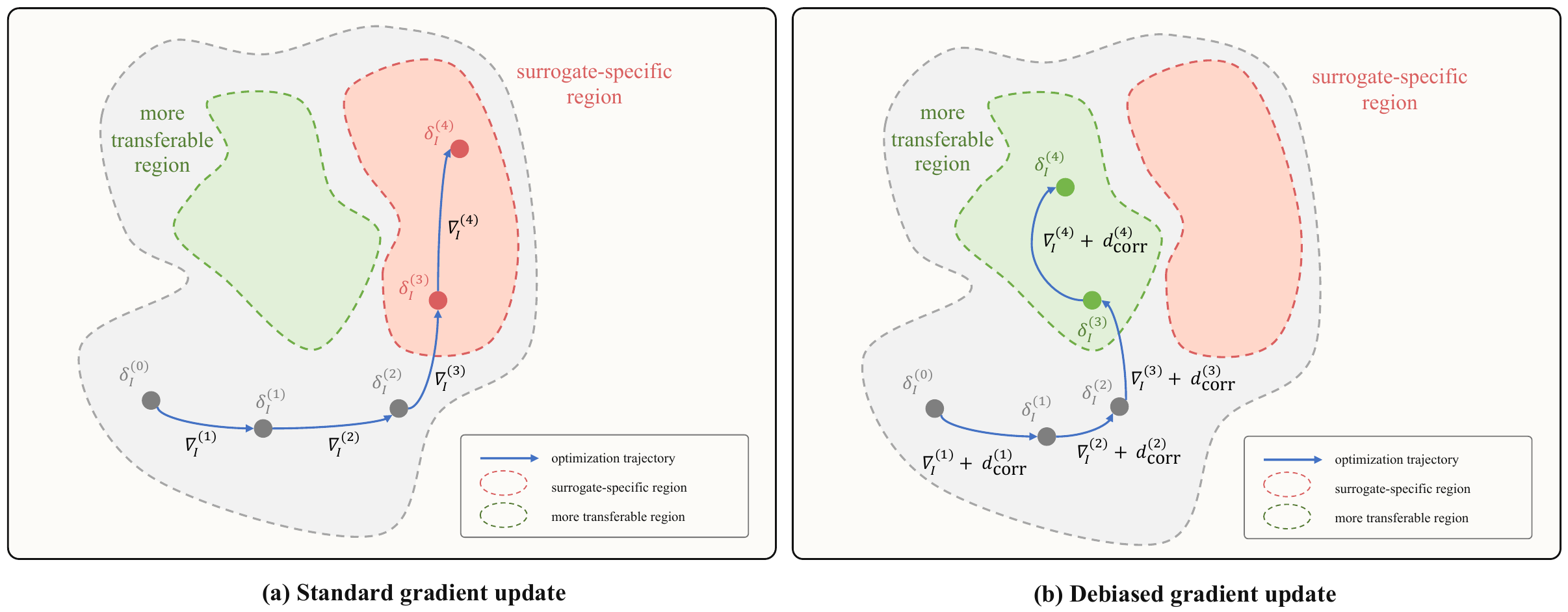}
    \vspace{-3pt}
    \caption{Conceptual comparison of gradient-update trajectories. (a) The standard gradient update used in previous methods directly follows the surrogate gradient and may drift toward surrogate-specific regions. (b) Our debiased gradient update suppresses bias-aligned components and steers the trajectory toward a more transferable region.}
    \label{fig:debias_concept}
\end{figure*}

To examine this dependency, we compare perturbations optimized with three visual inputs: the original image, Gaussian noise sampled at each iteration, and the dataset mean image. All three settings use the same surrogate misalignment objective, which maximizes the loss that decreases image-text similarity, and differ only in the visual input. The original image follows the standard attack process, where the perturbation depends on both input semantics and surrogate-model responses. The Gaussian noise is resampled at every iteration to remove meaningful visual semantics and disrupt fixed input structures. The dataset mean image preserves coarse dataset statistics while suppressing instance-level semantics. For the weak-semantic settings, the perturbation is generated on the weak input, but its transfer effect is evaluated after adding the resulting perturbation back to the clean image. These comparisons isolate how much adversarial optimization depends on surrogate-model tendencies.
Figure~\ref{fig:observation}(a)--(b) normalizes ASR by the CLIP$_{\text{ViT}}$ source result to compare relative transfer. Blue, orange, and green denote perturbations optimized from the original image, Gaussian noise, and the dataset mean image. All three inputs produce effective attacks on the CLIP$_{\text{ViT}}$ surrogate, reaching 100\% normalized ASR. After transfer, however, their behavior differs across targets. Perturbations optimized from weak-semantic inputs, including Gaussian noise and the dataset mean image, show larger drops on heterogeneous targets such as ALBEF and TCL, especially in text retrieval, while their transfer to the related CLIP$_{\text{CNN}}$ target remains stronger. This pattern suggests that adversarial optimization depends not only on the input itself; weak-semantic inputs can still produce effective surrogate attacks, but their transfer behavior is shaped by surrogate-model characteristics and target-model similarity.
We further measure gradient alignment to examine this transfer pattern. The cosine alignment between normalized surrogate and target input gradients reflects whether their local adversarial update directions are consistent. As shown in Figure~\ref{fig:observation}(c), perturbations optimized from the original image have higher alignment with heterogeneous targets such as ALBEF and TCL than those optimized from Gaussian noise or the dataset mean image. For the related CLIP$_{\text{CNN}}$ target, the alignment of weak-semantic inputs becomes higher than on heterogeneous targets, which is consistent with their stronger transfer to CLIP$_{\text{CNN}}$ in Figure~\ref{fig:observation}(a)--(b). These results suggest that adversarial update directions contain surrogate-specific bias across all optimization inputs. The gradient evidence further indicates that weak-semantic inputs carry stronger bias than the clean-image input: their update directions are less tied to transferable input semantics and more tied to surrogate-family behavior. We refer to this model-dependent component as surrogate-specific bias.

Based on this observation and the established connection between transferability and source-target gradient alignment~\cite{demontis2019adversarial,zhao2021success,zhao2023revisiting,gu2024a}, we propose DeBias-Attack, a transfer-based adversarial framework that reduces surrogate-specific bias during perturbation optimization. Figure~\ref{fig:debias_concept} compares the standard gradient update with our debiased update. Standard transfer-based attacks update perturbations by following surrogate-model gradients at each iteration. Although these directions increase the surrogate attack objective, some components may encode surrogate-specific behavior and transfer poorly to heterogeneous targets. As a result, an adversarial example can remain effective on the surrogate while losing strength after transfer.
Figure~\ref{fig:debias_concept}(b) illustrates our debiased gradient update. DeBias-Attack reduces surrogate-specific bias through bias-guided gradient correction. This update steers the perturbation trajectory away from surrogate-specific regions and toward directions with better cross-model transferability.

Specifically, DeBias-Attack maintains two image perturbation branches before the correction step. The main branch optimizes a perturbation on the original image to disrupt image-text alignment in the projected semantic subspace. To estimate surrogate-specific bias, the reference branch optimizes another perturbation on a weak-semantic image constructed from the dataset mean image by adding small Gaussian noise that is resampled at each iteration. The weak-semantic image preserves coarse dataset-level visual statistics but weakens clear instance-level visual semantics and fixed input structures. As a result, its optimization relies more on surrogate-model responses than on input semantics, so the resulting reference adversarial perturbation can be used to estimate surrogate-specific bias.
At each correction step, DeBias-Attack applies the main and reference perturbations to the original image and feeds the perturbed images into the surrogate model. The main perturbation produces the main adversarial gradient, while the reference perturbation produces a reference adversarial gradient reflecting surrogate-specific bias. DeBias-Attack reduces the positively aligned projection of the main gradient onto the reference gradient and uses the corrected direction to update the debiased adversarial image. Since this correction is performed throughout the iterative optimization, the whole perturbation trajectory is progressively debiased.
For the textual modality, DeBias-Attack performs context-aware token substitution after each image update. Candidate tokens are evaluated with the updated debiased adversarial image in the same projected semantic subspace, allowing the textual perturbation to follow the current image-side correction and further decrease image-text similarity. Through alternating image-text optimization, DeBias-Attack suppresses surrogate-specific bias and strengthens cross-modal semantic misalignment across models.

Experiments on multiple VLP models, downstream tasks, and multimodal large language models (MLLMs) validate the effectiveness of DeBias-Attack. Under black-box transfer settings, DeBias-Attack achieves competitive attack performance across different transfer scenarios, especially under heterogeneous transfer. These results show that bias-guided gradient correction improves the cross-model transferability of generated adversarial image-text pairs.

In summary, our work makes the following key contributions:
\begin{itemize}
    \item We show that adversarial update directions may depend on the surrogate model rather than only on the input semantics, which limits cross-model transferability across heterogeneous models. We refer to this dependency as surrogate-specific bias.
    \item We propose DeBias-Attack, which uses weak-semantic images to produce a reference adversarial gradient that reflects surrogate-specific bias, applies bias-guided gradient correction to reduce the influence of surrogate-specific bias during perturbation optimization, and refines textual perturbations with the updated debiased adversarial image.
    \item Experiments show that DeBias-Attack achieves strong black-box transfer performance across VLP models, downstream tasks, and MLLMs, including open-source and closed-source systems.
\end{itemize}

The remainder of this paper is organized as follows. Section~\ref{sec:related_work} reviews related work. Section~\ref{sec:threat_model} defines the threat model. Section~\ref{sec:the_proposed_method} presents DeBias-Attack. Section~\ref{sec:experiment} reports the experimental setup and results. Section~\ref{sec:conclusion} concludes the paper.

\section{Related Work}
\label{sec:related_work}

\subsection{VLP Models}

VLP models learn joint image-text representations from large-scale paired data~\cite{radford2021learning, li2022blip, li2020bert, dosovitskiy2020image}. Existing architectures encode and fuse modalities in different ways. Fusion-based models process multimodal inputs with shared transformers~\cite{chen2020uniter, tan2019lxmert}, dual-stream models use separate encoders with cross-modal interaction~\cite{lu2019vilbert, tan2019lxmert, li2021align, yang2022vision}, and dual-encoder models such as CLIP align separate image and text encodings in a shared space~\cite{radford2021learning, li2023blip}. This architectural diversity improves multimodal modeling but complicates transfer-based attacks, because perturbations optimized for one alignment mechanism may not generalize to heterogeneous targets. The gap among VLP architectures is a major obstacle for transferable attacks when the surrogate and target use different cross-modal alignment patterns.
Fusion-heavy models and contrastive dual-encoder models may respond to adversarial perturbations through different unimodal encoders, interaction modules, and embedding spaces. A perturbation that exploits one surrogate architecture can remain effective on related models but lose strength when transferred to another model family. This motivates studying transfer degradation through surrogate-specific optimization bias.

\subsection{Downstream Tasks of VLP Models}

VLP models are adapted to image-text retrieval, visual grounding, image captioning, and reasoning tasks. Retrieval evaluates cross-modal alignment by matching images and texts~\cite{li2021align}; grounding tests fine-grained region-text localization~\cite{yang2022improving, zhu2022seqtr, deng2021transvg}; captioning measures visual understanding and language generation~\cite{hossain2019comprehensive, zeng2024meacap}; and reasoning tasks assess higher-level multimodal inference~\cite{zellers2019recognition}. Since these tasks often reuse VLP backbones, transferable adversarial perturbations may affect multiple downstream systems. We also evaluate cross-task transfer from retrieval to localization and generation tasks built on related multimodal representations.
Among these tasks, retrieval is a common testbed for multimodal alignment because both image-to-text and text-to-image retrieval directly depend on the learned shared representation. Visual grounding and captioning provide complementary evidence: the former requires region-level correspondence between language and visual content, while the latter tests whether corrupted visual-textual alignment also affects generation. Evaluating these tasks together gives a more complete view of the practical risk posed by transferable VLP attacks.

\subsection{Adversarial Transferability}

Adversarial transferability allows perturbations to remain effective beyond the setting in which they are generated~\cite{gu2024a, wang2023beyond}. For VLP models, two forms are especially relevant: cross-task transfer, where attacks generated for one task affect other tasks~\cite{lu2020enhancing, zeng2024cross, jia2025semantic}, and cross-model transfer, where attacks crafted on a surrogate model remain effective on unseen target models~\cite{chen2024rethinking, fu2024improving}. Cross-model transfer can occur between related architectures or across heterogeneous models. Heterogeneous transfer is harder and more practical, because the attacker often does not know whether the victim model uses fusion, cross-attention, or contrastive alignment. Although robust training and defense methods have been explored, transferable multimodal attacks remain difficult to prevent~\cite{fares2024mirrorcheck}.
Cross-task transfer means one adversarial example can affect several downstream behaviors, while cross-model transfer enables black-box attacks without direct victim access. In VLP systems, these properties are connected because many downstream tasks inherit the same multimodal representations. Our work focuses on cross-model and cross-task transfer, with emphasis on heterogeneous model transfer.

Recent VLP attacks improve transferability through joint multimodal perturbation, set-level guidance, stronger optimization, semantic-aligned evolution, or global-local transformations~\cite{zhang2022towards, lu2023set, gao2024boosting, wang2024transferable, jia2025semantic, liu2025gleam}. These methods strengthen the attack objective or input transformation strategy. Our work instead studies transfer degradation caused by surrogate-specific bias and uses reference adversarial gradients from weak-semantic images to suppress this bias during perturbation optimization, which is useful for heterogeneous transfer.

\section{Threat Model}
\label{sec:threat_model}

\begin{figure*}[!t]
    \centering
    \includegraphics[width=0.90\linewidth]{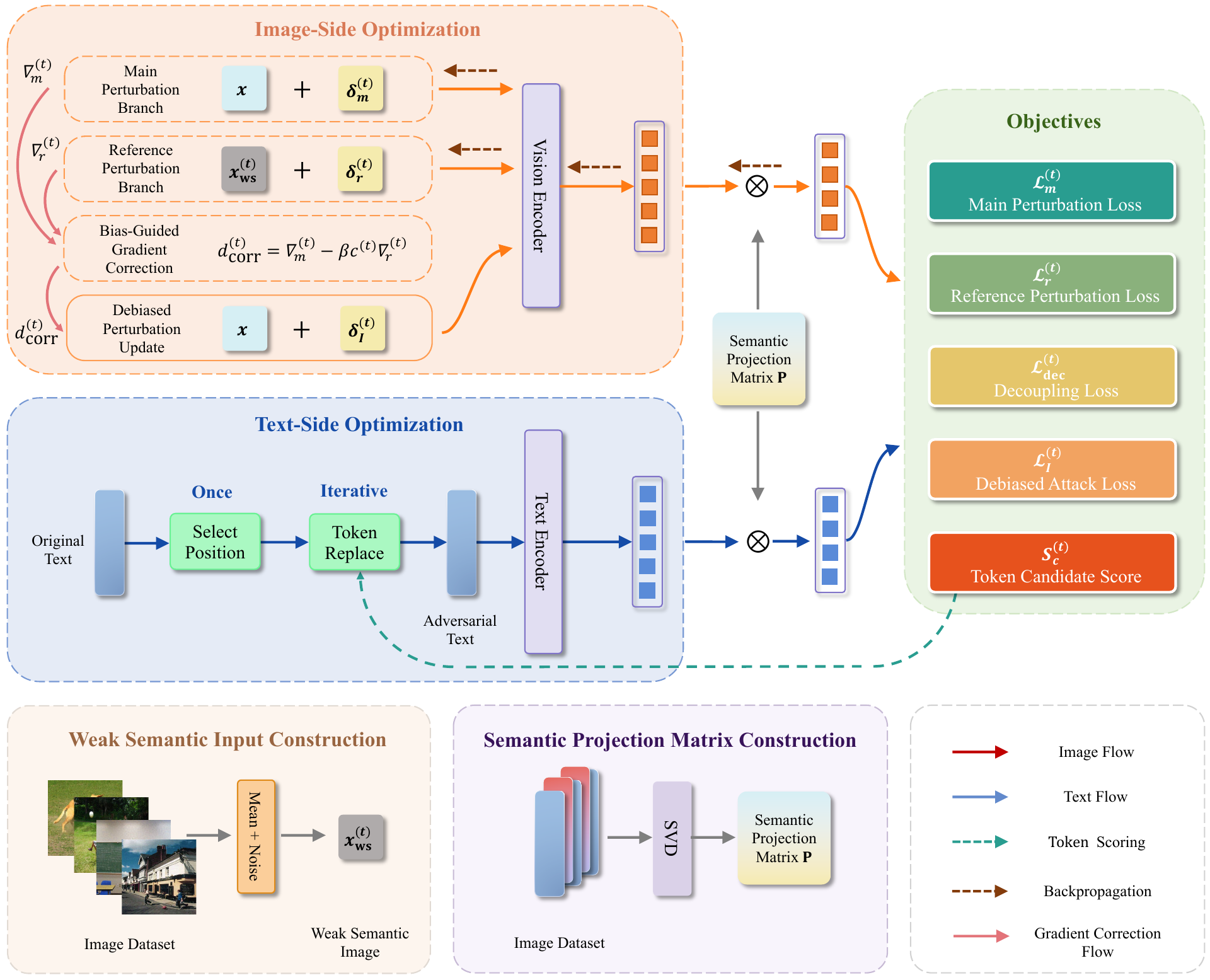}
    \vspace{-3pt}
    \caption{Overview of DeBias-Attack. The image-side optimization maintains main and reference perturbation branches, applies bias-guided gradient correction, and updates the debiased adversarial image. The text-side optimization selects token replacements according to the token candidate score to produce adversarial image-text pairs.}
    \label{fig:method}
\end{figure*}

\subsection{Attacker's goals.}
Adversarial objectives are commonly divided into targeted and untargeted settings. Targeted attacks force a specified output, whereas untargeted attacks degrade model performance without specifying the incorrect prediction. We focus on the untargeted setting and measure effectiveness by attack success rate and drops in task metrics.

\subsection{Attacker's constraints.}
We adopt a strict black-box, transfer-based setting. The attacker has benign image-text pairs and may use surrogate models for optimization, but has no access to the target model's architecture, parameters, gradients, outputs, or labels. Unlike query-based attacks, our setting allows no target interaction during generation or evaluation. Visual perturbations are bounded by a standard $\ell_p$ constraint to preserve imperceptibility, and textual perturbations are limited to one token substitution. The resulting adversarial image-text pairs are transferred to unseen targets, so all reported results reflect surrogate-only transferability.

\section{The Proposed Method}
\label{sec:the_proposed_method}

We first introduce the notation and problem setup, then describe adversarial image optimization and adversarial text optimization. DeBias-Attack maintains a main perturbation branch and a reference perturbation branch on a weak-semantic image, applies bias-guided gradient correction throughout the debiased perturbation update, and optimizes the adversarial text with the updated debiased adversarial image. The supplementary material gives the theoretical analysis.

\begin{algorithm}[t]
\caption{Optimization Procedure of DeBias-Attack}
\label{alg:debias}
\small
\begin{algorithmic}[1]
\STATE \textbf{Input:} $f_I,f_T,\mathbf{P},(\mathbf{x}_I,\mathbf{x}_T)$
\STATE \textbf{Output:} $\tilde{\mathbf{x}}_I^{T}, \tilde{\mathbf{x}}_T^{T}$

\STATE Initialize $\boldsymbol{\delta}_{m}^{0}$, $\boldsymbol{\delta}_{r}^{0}$, $\boldsymbol{\delta}_{I}^{0}$ within $\varepsilon$
\STATE $\tilde{\mathbf{x}}_I^{0}\gets\Pi_{[0,1]}(\mathbf{x}_I+\boldsymbol{\delta}_{I}^{0})$, $\tilde{\mathbf{x}}_T^{0}\gets\mathbf{x}_T$

\STATE \textbf{Select token position by masking}
\FOR{each token position $i$ in $\tilde{\mathbf{x}}_T^{0}$}
    \STATE Construct masked text $\mathbf{x}_T^{\setminus i}$
    \STATE $\Delta_i \gets l(\mathbf{x}_I,\mathbf{x}_T^{\setminus i})+
    l(\tilde{\mathbf{x}}_I^{0},\mathbf{x}_T^{\setminus i})$
\ENDFOR
\STATE $i^\star \gets \arg\max_i \Delta_i$

\FOR{$t=0$ to $T-1$}
\STATE \textbf{Main branch}
\STATE Optimize the main perturbation on the original image
\STATE $\mathcal{L}_{m}^{(t)}
    =l(\mathbf{x}_I+\boldsymbol{\delta}_{m}^{(t)},\tilde{\mathbf{x}}_T^{(t)})$
\STATE $\boldsymbol{\delta}_{m}^{(t+1)}
    \gets
    \Pi_{\varepsilon}
    \left(
    \boldsymbol{\delta}_{m}^{(t)}
    +
    \alpha\,\mathrm{sign}
    \left(
    \nabla_{\boldsymbol{\delta}_{m}^{(t)}}\mathcal{L}_{m}^{(t)}
    \right)
    \right)$

\STATE \textbf{Reference branch}
\STATE Construct a weak-semantic image with resampled Gaussian noise
\STATE Sample $\boldsymbol{\xi}^{(t)}\sim\mathcal{N}(0,\rho^2)$
\STATE $\mathbf{x}_{\mathrm{ws}}^{(t)}
    \gets
    \Pi_{[0,1]}
    \left(
    \frac{1}{N}\sum_{n=1}^{N}\mathbf{x}^{(n)}
    +
    \boldsymbol{\xi}^{(t)}
    \right)$
\STATE Optimize the reference perturbation and decouple it from the main branch
\STATE $\mathcal{L}_{r}^{(t)}
    =l(\mathbf{x}_{\mathrm{ws}}^{(t)}+\boldsymbol{\delta}_{r}^{(t)},\tilde{\mathbf{x}}_T^{(t)})$
\STATE $\mathcal{L}_{\mathrm{dec}}^{(t)}
    =
    -\cos\!\left(
    f_I(\mathbf{x}_{I}+\boldsymbol{\delta}_{m}^{(t)})\mathbf{P},
    f_I(\mathbf{x}_{I}+\boldsymbol{\delta}_{r}^{(t)})\mathbf{P}
    \right)$
\STATE $\boldsymbol{\delta}_{r}^{(t+1)}
    \gets
    \Pi_{\varepsilon}
    \left(
    \boldsymbol{\delta}_{r}^{(t)}
    +
    \alpha\,\mathrm{sign}
    \left(
    \nabla_{\boldsymbol{\delta}_{r}^{(t)}}
    \left(
    \mathcal{L}_{r}^{(t)}
    +
    \mathcal{L}_{\mathrm{dec}}^{(t)}
    \right)
    \right)
    \right)$

\STATE \textbf{Bias-guided correction}
\STATE Compute main and reference gradients under the original-image context
\STATE $\nabla_m^{(t)}
    =
    \nabla_{\boldsymbol{\delta}_{m}^{(t)}}
    l(\mathbf{x}_{I}+\boldsymbol{\delta}_{m}^{(t)},\tilde{\mathbf{x}}_T^{(t)})$
\STATE $\nabla_r^{(t)}
    =
    \nabla_{\boldsymbol{\delta}_{r}^{(t)}}
    l(\mathbf{x}_{I}+\boldsymbol{\delta}_{r}^{(t)},\tilde{\mathbf{x}}_T^{(t)})$
\STATE Remove the positively aligned reference-gradient component
\STATE $c^{(t)}
    \gets
    \dfrac{
    \max(\langle\nabla_m^{(t)},\nabla_r^{(t)}\rangle,0)
    }{
    \|\nabla_r^{(t)}\|_2^2+\epsilon_{\mathrm{num}}
    }$
\STATE $\mathbf{d}_{\mathrm{corr}}^{(t)}
    \gets
    \nabla_m^{(t)}
    -
    \beta c^{(t)}\nabla_r^{(t)}$

\STATE \textbf{Debiased update}
\STATE Update the final image perturbation with the corrected direction
\STATE $\mathcal{L}_{I}^{(t)}
    =
    l(\mathbf{x}_I+\boldsymbol{\delta}_{I}^{(t)},\tilde{\mathbf{x}}_T^{(t)})$
\STATE $\boldsymbol{\delta}_{I}^{(t+1)}
    \gets
    \Pi_{\varepsilon}
    \left(
    \boldsymbol{\delta}_{I}^{(t)}
    +
    \alpha\,\mathrm{sign}
    \left(
    \nabla_{\boldsymbol{\delta}_{I}^{(t)}}\mathcal{L}_{I}^{(t)}
    +
    \mathbf{d}_{\mathrm{corr}}^{(t)}
    \right)
    \right)$
\STATE $\tilde{\mathbf{x}}_I^{(t+1)}
    \gets
    \Pi_{[0,1]}(\mathbf{x}_I+\boldsymbol{\delta}_{I}^{(t+1)})$

\STATE \textbf{Text update}
\STATE Generate candidates at the selected position and score them with the updated image
\STATE $\mathcal{C}_{i^\star}^{(t)}
    \gets
    \mathrm{MLM}(\tilde{\mathbf{x}}_T^{(t)},i^\star,C)$
\FOR{each candidate token $c\in\mathcal{C}_{i^\star}^{(t)}$}
        \STATE Construct $\mathbf{x}_T^{(c,t)}$ by replacing $i^\star$ with $c$
        \STATE $S_c^{(t)}
        \gets
        l(\mathbf{x}_I,\mathbf{x}_T^{(c,t)})
        +
        l(\tilde{\mathbf{x}}_I^{(t+1)},\mathbf{x}_T^{(c,t)})$
\ENDFOR
\STATE $c_t^\star \gets \arg\max_{c\in\mathcal{C}_{i^\star}^{(t)}} S_c^{(t)}$
\STATE $\tilde{\mathbf{x}}_T^{(t+1)}\gets\mathbf{x}_T^{(c_t^\star,t)}$
\ENDFOR

\STATE \textbf{return} $\tilde{\mathbf{x}}_I^{T}, \tilde{\mathbf{x}}_T^{T}$
\end{algorithmic}
\end{algorithm}

DeBias-Attack reduces surrogate dependency through bias-guided gradient correction guided by weak-semantic images.
Figure~\ref{fig:method} shows the framework. Visual and textual representations are evaluated in a projected semantic subspace, where cosine similarity measures image-text alignment. On the image side, the main branch optimizes a perturbation on the original image with the main perturbation loss. The reference branch optimizes a perturbation on a randomized weak-semantic image with the reference perturbation loss and the decoupling loss. Both perturbations are added to the original image to compute the main and reference adversarial gradients for bias-guided correction. The debiased perturbation update combines the corrected direction with the debiased attack loss to generate the debiased adversarial image. On the text side, token substitutions are refined with the updated debiased adversarial image according to the token candidate score, so visual and textual perturbations reduce cross-modal semantic similarity together.

\subsection{Notation and Problem Setup}

Let $\mathbf{x}_{I}\in[0,1]^{C\times H\times W}$ denote the input image and $\mathbf{x}_{T}=(t_1,t_2,\ldots,t_L)\in\mathcal{V}^{L}$ denote the input text, where $t_i\in\mathcal{V}$ is the $i$-th token, $\mathcal{V}$ is the vocabulary, and $L$ is the sequence length. The vision encoder $f_I(\cdot)$ maps the image to a visual representation $\mathbf{z}_{I}=f_I(\mathbf{x}_{I})$, and the text encoder $f_T(\cdot)$ maps the text to a textual representation $\mathbf{z}_{T}=f_T(\mathbf{x}_{T})$. Given the original image-text pair $(\mathbf{x}_{I},\mathbf{x}_{T})$ and the surrogate encoders $(f_I,f_T)$, our goal is to construct an adversarial image $\tilde{\mathbf{x}}_I$ and an adversarial text $\tilde{\mathbf{x}}_T$ that disrupt image-text alignment while keeping the image perturbation imperceptible and the textual perturbation fluent and semantically plausible.

For the visual modality, the adversarial image is obtained by adding a bounded perturbation $\boldsymbol{\delta}_I$ to the original image:
\begin{equation}
    \tilde{\mathbf{x}}_{I}
    =
    \Pi_{[0,1]}\!\left(\mathbf{x}_{I}+\boldsymbol{\delta}_I\right),
    \qquad
    \|\boldsymbol{\delta}_I\|_{\infty}\le \varepsilon,
    \label{eq:adv_image}
\end{equation}
where $\Pi_{[0,1]}(\cdot)$ clips the adversarial image into the valid pixel range, and the $\ell_\infty$ constraint keeps the image perturbation within the attack budget. In the following image-side optimization, $\boldsymbol{\delta}_I$ denotes the debiased adversarial perturbation produced by the proposed correction strategy.

For the textual modality, the adversarial text is obtained by replacing a small subset of tokens in the original text:
\begin{equation}
    \tilde{\mathbf{x}}_{T}
    =
    (\tilde{t}_1,\tilde{t}_2,\ldots,\tilde{t}_L)
    \in \mathcal{V}^{L},
    \qquad
    d_H(\tilde{\mathbf{x}}_T,\mathbf{x}_T)\le \varepsilon_T,
    \label{eq:adv_text}
\end{equation}
where $\tilde{t}_i$ denotes the substituted token at position $i$ and $d_H(\cdot,\cdot)$ measures the number of replaced token positions. This constraint preserves fluency and semantic plausibility by limiting the total number of token substitutions.

With the adversarial image and text defined above, image-text alignment is measured by the cosine similarity between visual and textual representations:
\begin{equation}
\begin{aligned}
s(\mathbf{x}_{I},\mathbf{x}_{T})
&=
\cos\!\left(f_I(\mathbf{x}_{I}),f_T(\mathbf{x}_{T})\right) \\
&=
\frac{
f_I(\mathbf{x}_{I}) f_T(\mathbf{x}_{T})^{\top}
}{
\|f_I(\mathbf{x}_{I})\|_2
\|f_T(\mathbf{x}_{T})\|_2
}.
\end{aligned}
\label{eq:cosine_similarity}
\end{equation}

Under the image and text perturbation constraints in Eq.~\eqref{eq:adv_image} and Eq.~\eqref{eq:adv_text}, the adversarial objective is to reduce the image-text similarity of the adversarial pair, which is equivalently formulated as maximizing the corresponding misalignment score:
\begin{equation}
\begin{aligned}
\min_{\tilde{\mathbf{x}}_I,\tilde{\mathbf{x}}_T}
\quad
&s(\tilde{\mathbf{x}}_I,\tilde{\mathbf{x}}_T)
\\
\Longleftrightarrow
\quad
\max_{\tilde{\mathbf{x}}_I,\tilde{\mathbf{x}}_T}
\quad
&\left[-s(\tilde{\mathbf{x}}_I,\tilde{\mathbf{x}}_T)\right].
\end{aligned}
\label{eq:attack_goal}
\end{equation}

\subsection{Adversarial Image Optimization}

DeBias-Attack optimizes the image perturbation in four steps: optimizing the main perturbation branch, optimizing the reference perturbation branch, performing bias-guided gradient correction, and updating the debiased perturbation. The main branch optimizes the adversarial perturbation on the original image and provides a gradient that may contain both transferable image-text misalignment directions and surrogate-specific directions. The reference branch learns from randomized weak-semantic images and provides a gradient used to estimate surrogate-specific bias. Bias-guided gradient correction reduces the part of the main gradient that aligns with the reference gradient. The debiased perturbation update applies the corrected direction to $\boldsymbol{\delta}_I$ while preserving attack strength.

Following SA-AET~\cite{jia2025semantic}, we construct a semantic projection matrix $\mathbf{P}\in\mathbb{R}^{d\times d}$ from text embeddings in the Attack Text Corpus. We build this corpus by sampling 40\% of the text instances from the dataset, using the same setting as SA-AET. The sampling is independent of the attack objective and does not use adversarial gradients. We apply singular value decomposition to the sampled corpus embeddings to obtain $\mathbf{P}$. The projection compresses unstable dimensions and dimensions with limited semantic information, so image and text features are compared in a concentrated semantic subspace. The projection matrix is not adapted to any target model. For an image-text pair $(\mathbf{x}_I,\mathbf{x}_T)$, we define the projected cosine similarity as:
\begin{equation}
    s_{\mathbf{P}}(\mathbf{x}_I,\mathbf{x}_T)
    =
    \cos\!\left(
    f_I(\mathbf{x}_I)\mathbf{P},
    f_T(\mathbf{x}_T)\mathbf{P}
    \right).
    \label{eq:projected_similarity}
\end{equation}
The image-side attack decreases this projected similarity. We maximize the misalignment score:
\begin{equation}
    l(\mathbf{x}_I,\mathbf{x}_T)
    =
    -s_{\mathbf{P}}(\mathbf{x}_I,\mathbf{x}_T).
    \label{eq:misalignment_score}
\end{equation}
This objective pushes visual and textual representations apart in the projected semantic subspace.

\textbf{Main perturbation branch.}
Let $\boldsymbol{\delta}_{m}$ denote the auxiliary main adversarial perturbation. At iteration $t$, it is optimized on the original image:
\begin{equation}
    \mathcal{L}_{m}^{(t)}
    =
    l\!\left(
    \mathbf{x}_{I}
    +
    \boldsymbol{\delta}_{m}^{(t)},
    \tilde{\mathbf{x}}_{T}^{(t)}
    \right).
    \label{eq:main_loss}
\end{equation}
Maximizing the main perturbation loss $\mathcal{L}_{m}^{(t)}$ decreases the projected cosine similarity between the perturbed image and the current text. The main adversarial perturbation is updated as:
\begin{equation}
    \boldsymbol{\delta}_{m}^{(t+1)}
    =
    \Pi_{\varepsilon}
    \left(
    \boldsymbol{\delta}_{m}^{(t)}
    +
    \alpha\,
    \mathrm{sign}\!\left(
    \nabla_{\boldsymbol{\delta}_{m}^{(t)}}
    \mathcal{L}_{m}^{(t)}
    \right)
    \right),
    \label{eq:main_update}
\end{equation}
where $\alpha$ is the common step size for image-side perturbation updates.

\textbf{Reference perturbation branch.}
DeBias-Attack constructs a randomized weak-semantic image at each iteration to estimate surrogate-specific bias. It adds small Gaussian noise to the dataset mean image, preserving coarse dataset-level image statistics while weakening stable visual semantics and fixed image structures. The Gaussian noise is re-sampled at each iteration. Given the dataset $\mathcal{X}=\{\mathbf{x}^{(n)}\}_{n=1}^{N}$, the weak-semantic image is defined as:
\begin{equation}
    \mathbf{x}_{\mathrm{ws}}^{(t)}
    =
    \Pi_{[0,1]}\!\left(
    \frac{1}{N}
    \sum_{n=1}^{N}
    \mathbf{x}^{(n)}
    +
    \boldsymbol{\xi}^{(t)}
    \right),
    \label{eq:weak_semantic_input}
\end{equation}
where $\boldsymbol{\xi}^{(t)}$ is re-sampled from a Gaussian distribution $\mathcal{N}(0,\rho^2)$ at each iteration, and $\rho$ controls the standard deviation of the random perturbation added to the dataset mean image. The reference adversarial perturbation $\boldsymbol{\delta}_{r}$ is optimized on this weak-semantic image paired with the current adversarial text $\tilde{\mathbf{x}}_{T}^{(t)}$:
\begin{equation}
    \mathcal{L}_{r}^{(t)}
    =
    l\!\left(
    \mathbf{x}_{\mathrm{ws}}^{(t)}
    +
    \boldsymbol{\delta}_{r}^{(t)},
    \tilde{\mathbf{x}}_{T}^{(t)}
    \right).
    \label{eq:reference_loss}
\end{equation}
Because the weak-semantic image comes from the dataset mean plus Gaussian noise, it reduces dependence on clear visual semantics and fixed input structures. The reference perturbation therefore reflects surrogate-specific bias rather than image-specific semantics.

The reference perturbation should not copy the main perturbation trajectory. We apply both perturbations to the same original image and reduce the similarity between their projected visual representations with the decoupling loss:
\begin{equation}
\mathcal{L}_{\mathrm{dec}}^{(t)}
=
-\cos\!\left(
    f_I(\mathbf{x}_{I}+\boldsymbol{\delta}_{m}^{(t)})\mathbf{P},
    f_I(\mathbf{x}_{I}+\boldsymbol{\delta}_{r}^{(t)})\mathbf{P}
\right).
\label{eq:decoupling_loss}
\end{equation}
Maximizing this term reduces the similarity between the projected visual responses induced by the current main and reference perturbations. This moves the reference perturbation away from the original attack direction and makes its gradient more suitable for correction.

Combining the reference perturbation loss with the decoupling loss, the reference adversarial perturbation is updated as:
\begin{equation}
    \boldsymbol{\delta}_{r}^{(t+1)}
    =
    \Pi_{\varepsilon}
    \left(
    \boldsymbol{\delta}_{r}^{(t)}
    +
    \alpha\,
    \mathrm{sign}\!\left(
    \nabla_{\boldsymbol{\delta}_{r}^{(t)}}
    \left(
    \mathcal{L}_{r}^{(t)}
    +
    \mathcal{L}_{\mathrm{dec}}^{(t)}
    \right)
    \right)
    \right).
    \label{eq:reference_update}
\end{equation}

\textbf{Bias-guided gradient correction.}
At iteration $t$, DeBias-Attack applies the current main and reference perturbations to the original image and feeds the perturbed images into the surrogate model. The main perturbation produces the gradient used to disrupt image-text alignment. The reference perturbation, learned from weak-semantic images, encodes surrogate-preferred perturbation patterns.
For the correction step, we compute two gradients under the same original-image context:
\begin{equation}
\begin{aligned}
    \nabla_m^{(t)}
    &=
    \nabla_{\boldsymbol{\delta}_{m}^{(t)}}
    l\!\left(
    \mathbf{x}_{I}
    +
    \boldsymbol{\delta}_{m}^{(t)},
    \tilde{\mathbf{x}}_{T}^{(t)}
    \right),\\
    \nabla_r^{(t)}
    &=
    \nabla_{\boldsymbol{\delta}_{r}^{(t)}}
    l\!\left(
    \mathbf{x}_{I}
    +
    \boldsymbol{\delta}_{r}^{(t)},
    \tilde{\mathbf{x}}_{T}^{(t)}
    \right),
\end{aligned}
\label{eq:main_reference_gradients}
\end{equation}
where $\nabla_m^{(t)}$ is the main gradient induced by applying the main perturbation to the original image, while $\nabla_r^{(t)}$ is the reference gradient induced by applying the reference perturbation to the same image. Computing both gradients under the same visual context makes their projection relationship well defined and lets the reference gradient indicate which part of the main gradient follows surrogate-specific bias.

We compute the positive projection coefficient from the main adversarial gradient onto the reference adversarial gradient:
\begin{equation}
    c^{(t)}
    =
    \frac{
    \max\!\left(
    \left\langle
    \nabla_m^{(t)},
    \nabla_r^{(t)}
    \right\rangle,
    0
    \right)
    }{
    \left\|
    \nabla_r^{(t)}
    \right\|_{2}^{2}
    +
    \epsilon_{\mathrm{num}}
    },
    \label{eq:projection_coeff}
\end{equation}
where $\epsilon_{\mathrm{num}}$ is a small numerical constant used to avoid division by zero. All inner products and norms are computed after vectorizing the gradients. The coefficient $c^{(t)}$ measures how much of the main gradient aligns with the reference gradient. The maximum operator applies the correction only when the two gradients are positively aligned.

The corrected direction reduces the bias-related part of the main gradient:
\begin{equation}
    \mathbf{d}_{\mathrm{corr}}^{(t)}
    =
    \nabla_m^{(t)}
    -
    \beta c^{(t)}
    \nabla_r^{(t)} ,
    \label{eq:corrected_direction}
\end{equation}
where $\beta$ controls the correction strength. Since $c^{(t)}$ is computed from the positive projection coefficient in Eq.~\eqref{eq:projection_coeff}, the correction removes only the positively aligned component of the main gradient along the reference gradient.

\textbf{Debiased perturbation update.}
The debiased adversarial perturbation $\boldsymbol{\delta}_{I}$ is optimized on the original image with the debiased attack loss to preserve adversarial effectiveness:
\begin{equation}
    \mathcal{L}_{I}^{(t)}
    =
    l\!\left(
    \mathbf{x}_{I}
    +
    \boldsymbol{\delta}_{I}^{(t)},
    \tilde{\mathbf{x}}_{T}^{(t)}
    \right).
    \label{eq:debiased_attack_loss}
\end{equation}
During optimization, the debiased adversarial perturbation is updated by combining its own adversarial gradient with the corrected direction:
\begin{equation}
    \boldsymbol{\delta}_{I}^{(t+1)}
    =
    \Pi_{\varepsilon}
    \left(
    \boldsymbol{\delta}_{I}^{(t)}
    +
    \alpha\,
    \mathrm{sign}\!\left(
    \nabla_{\boldsymbol{\delta}_{I}^{(t)}}
    \mathcal{L}_{I}^{(t)}
    +
    \mathbf{d}_{\mathrm{corr}}^{(t)}
    \right)
    \right).
    \label{eq:final_update}
\end{equation}
The debiased adversarial image is formed as:
\begin{equation}
    \tilde{\mathbf{x}}_{I}^{(t+1)}
    =
    \Pi_{[0,1]}\!\left(
    \mathbf{x}_{I}
    +
    \boldsymbol{\delta}_{I}^{(t+1)}
    \right).
    \label{eq:image_generation}
\end{equation}
The main and reference branches provide the gradients used for correction, while the debiased update maintains $\boldsymbol{\delta}_{I}$ as the final adversarial perturbation. This separated update preserves attack strength while reducing the influence of surrogate-specific bias, improving cross-model transferability.

\subsection{Adversarial Text Optimization}

To perturb the textual modality while preserving fluency and semantic plausibility, we use a gradient-free substitution strategy. The text update follows each image-side correction step. Candidate substitutions are evaluated with the current debiased adversarial image under the same projected misalignment score, keeping the text perturbation synchronized with the image update.

Before alternating optimization, we select the token position to attack. For each token position $i$, we construct a masked text $\mathbf{x}_{T}^{\setminus i}$ by masking the $i$-th token. We measure token importance under both the original image and the initial adversarial image:
\begin{equation}
    i^\star
    =
    \arg\max_i \Delta_i,
    \qquad
    \Delta_i
    =
    l(\mathbf{x}_I,\mathbf{x}_{T}^{\setminus i})
    +
    l(\tilde{\mathbf{x}}_I^{(0)},\mathbf{x}_{T}^{\setminus i}).
    \label{eq:token_importance}
\end{equation}
This score identifies the token whose masking increases projected image-text misalignment most under the original and initial adversarial visual contexts. The position search is performed once to reduce cost, and the selected position $i^\star$ is kept fixed during subsequent iterations.
After fixing the attack position $i^\star$, we use a pretrained BertForMaskedLM model to generate context-aware candidate substitutions. Let $\mathcal{C}_{i^\star}^{(t)}$ denote the candidate set generated for the current adversarial text $\tilde{\mathbf{x}}_T^{(t)}$. For each candidate token $c\in\mathcal{C}_{i^\star}^{(t)}$, we construct a candidate text $\mathbf{x}_{T}^{(c,t)}$ by replacing the token at position $i^\star$ with $c$. The token candidate score is computed under both the original image and the updated debiased adversarial image:
\begin{equation}
\begin{aligned}
    S_c^{(t)}
    =
    l(\mathbf{x}_I,\mathbf{x}_{T}^{(c,t)})
    +
    l(\tilde{\mathbf{x}}_I^{(t+1)},\mathbf{x}_{T}^{(c,t)}),\\
    c_t^\star
    =
    \arg\max_{c\in\mathcal{C}_{i^\star}^{(t)}} S_c^{(t)} .
\end{aligned}
    \label{eq:text_candidate}
\end{equation}
The adversarial text is updated as $\tilde{\mathbf{x}}_T^{(t+1)}=\mathbf{x}_{T}^{(c_t^\star,t)}$. Because the candidate token is evaluated with the updated debiased adversarial image, the textual perturbation follows the current image-side correction result. This alternating image-text update decreases projected image-text similarity and produces adversarial image-text pairs with stronger cross-model transferability. Algorithm~\ref{alg:debias} summarizes the full procedure.

\section{Experiments}
\label{sec:experiment}

We evaluate DeBias-Attack across cross-model transfer, cross-task transfer, perturbation ablations, and transfer to multimodal large language models (MLLMs). These experiments test attack effectiveness from complementary perspectives: transfer across architectures, transfer across downstream objectives, sensitivity to the debiasing design, and transfer to open-ended multimodal generation. The supplementary material includes additional MLLM qualitative examples, attention maps, modality ablations, and defense evaluation.

\subsection{Experiment Setup}

\subsubsection{Datasets}

We use Flickr30K~\cite{plummer2015flickr30k} and MSCOCO~\cite{lin2014microsoft} for image-text retrieval, RefCOCO+~\cite{yu2016modeling} for visual grounding, and MSCOCO for image captioning. These datasets cover retrieval, localization, and generation scenarios, enabling evaluation of transferable multimodal adversarial examples across different downstream tasks.

\subsubsection{Surrogate and Target Models}

We use ALBEF~\cite{li2021align}, TCL~\cite{yang2022vision}, CLIP$_{\rm CNN}$~\cite{radford2021learning}, and CLIP$_{\rm ViT}$~\cite{radford2021learning} as surrogate and target models. ALBEF and TCL involve tighter multimodal interaction, while CLIP$_{\rm CNN}$ and CLIP$_{\rm ViT}$ follow contrastive image-text alignment with different visual backbones. This model set covers both related and heterogeneous transfer settings and tests whether an attack remains effective when the target alignment mechanism differs from the surrogate.

\subsubsection{Downstream Tasks}

We evaluate Image-Text Retrieval (ITR), Visual Grounding (VG), and Image Captioning (IC). ITR is the attack generation task, while VG and IC test transfer to localization and generation without additional optimization.

\subsubsection{Evaluation Metrics}

Following prior work, we use Attack Success Rate (ASR) as the primary metric and compute it on original success, so the evaluation focuses on samples that the clean model handles correctly. For retrieval, we report R@1 ASR for text retrieval (TR, image-to-text retrieval) and image retrieval (IR, text-to-image retrieval), where rank$_\text{original}=1$ and rank$_\text{adv}>1$ indicates success. For captioning, we report BLEU~\cite{papineni2002bleu}, METEOR~\cite{banerjee2005meteor}, ROUGE~\cite{lin2004rouge}, CIDEr~\cite{vedantam2015cider}, and SPICE~\cite{anderson2016spice}; for grounding, we report localization accuracy under IoU $\ge$ 0.5.

\subsubsection{Baseline Methods}

We compare with PGD~\cite{madry2017towards}, BERT-Attack~\cite{li2020bert}, Co-Attack~\cite{zhang2022towards}, SGA~\cite{lu2023set}, DRA~\cite{gao2024boosting}, SA-AET~\cite{jia2025semantic}, GLEAM~\cite{liu2025gleam}, and TMM~\cite{wang2024transferable}. Together, these methods cover attacks on images, text, embeddings, multimodal perturbations, trajectory optimization, semantic evolution, and input transformations.

\subsubsection{Implementation Details}

All experiments are implemented in PyTorch on an NVIDIA RTX~3090 GPU. Unless otherwise specified, image-side optimization uses 8 iterations, perturbation budget $\varepsilon=8/255$, and step size $\alpha=2/255$ for the main, reference, and debiased adversarial perturbations. We set the Gaussian noise standard deviation to $\rho=2/255$ and set the correction strength to $\beta=0.9$.

\subsection{Qualitative Attack Examples}

\begin{figure}[!t]
    \centering
    \includegraphics[width=\columnwidth]{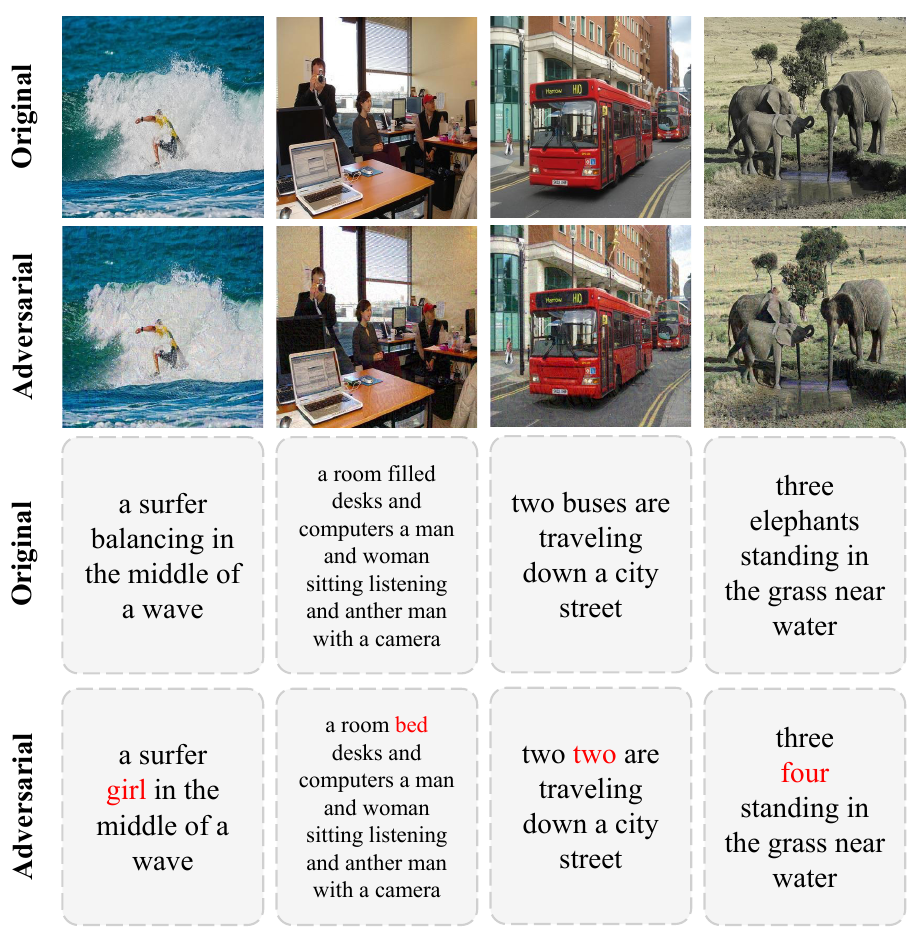}
    \vspace{-3pt}
    \caption{Representative adversarial examples. Each case compares the original and adversarial behavior; incorrect or mismatched retrieval results indicate successful attacks.}
    \label{fig:examples}
\end{figure}

Figure~\ref{fig:examples} shows representative adversarial examples generated by DeBias-Attack. Although the visual perturbations remain subtle, the adversarial image-text pairs lead to clear retrieval failures across heterogeneous target models. The generated perturbations cause cross-modal semantic misalignment beyond the surrogate model. This qualitative experiment provides a direct view of the attack effect.

\subsection{Cross-Model Attack Performance Evaluation}

\begin{table*}[!t]
\caption{R@1 attack success rates (\%) on Flickr30K for image and text retrieval. ``Surrogate'' denotes the VLP model used to generate adversarial examples; higher values indicate stronger attacks, and bold values mark the best result for each surrogate-target retrieval setting (ties included).}
\label{tab:cross_model_flickr}
\centering
\footnotesize
\renewcommand{\arraystretch}{1}
\setlength{\tabcolsep}{8pt}
\resizebox{0.95\textwidth}{!}{
\begin{tabular}{@{\extracolsep{\fill}} c|c|cc|cc|cc|cc}
\toprule[0.3mm]
& \textbf{Target} & \multicolumn{2}{c}{\textbf{ALBEF}} & \multicolumn{2}{c}{\textbf{TCL}} & \multicolumn{2}{c}{\textbf{CLIP$_{\rm ViT}$}} & \multicolumn{2}{c}{\textbf{CLIP$_{\rm CNN}$}} \\
\cmidrule{2-10}
\multirow{-2}{*}{\textbf{Surrogate}} & \textbf{Attack} & TR R@1 & IR R@1 & TR R@1 & IR R@1 & TR R@1 & IR R@1 & TR R@1 & IR R@1 \\
\midrule

\multirow{9}{*}{\textbf{ALBEF}}
& PGD~\cite{madry2017towards} & 92.21 & 93.88 & 24.47 & 28.36 & 10.95 & 16.01 & 14.32 & 19.42 \\
& BERT-Attack~\cite{li2020bert} & 11.89 & 27.11 & 12.21 & 27.65 & 29.61 & 43.45 & 32.37 & 45.79 \\
& Co-Attack~\cite{zhang2022towards} & 96.72 & 98.12 & 39.88 & 50.63 & 30.03 & 39.42 & 31.70 & 42.31 \\
& SGA~\cite{lu2023set} & 99.86 & 99.95 & 87.42 & 88.56 & 37.05 & 47.01 & 39.91 & 49.53 \\
& DRA~\cite{gao2024boosting} & 99.83 & 99.92 & 91.88 & 90.96 & 46.73 & 56.25 & 49.92 & 59.24 \\
& TMM~\cite{wang2024transferable} & 97.53 & 97.51 & 64.97 & 69.60 & 52.90 & 60.90 & 56.61 & 62.97 \\
& SA-AET~\cite{jia2025semantic} & 99.91 & 99.97 & 96.20 & 95.79 & 55.84 & 63.55 & 57.63 & 65.28 \\
& GLEAM~\cite{liu2025gleam} & \textbf{100.0} & 99.97 & 96.27 & 95.91 & 66.92 & 66.19 & 67.09 & 66.26 \\
& \textbf{DeBias-Attack} & \textbf{100.0} & \textbf{100.0} & \textbf{97.68} & \textbf{97.33} & \textbf{70.06} & \textbf{73.26} & \textbf{70.88} & \textbf{74.96} \\
\midrule

\multirow{9}{*}{\textbf{TCL}}
& PGD~\cite{madry2017towards} & 35.10 & 41.28 & 89.08 & 89.45 & 10.33 & 16.58 & 14.57 & 20.96 \\
& BERT-Attack~\cite{li2020bert} & 12.04 & 27.02 & 14.90 & 29.36 & 29.41 & 43.88 & 33.61 & 45.71 \\
& Co-Attack~\cite{zhang2022towards} & 50.20 & 59.74 & 91.29 & 95.10 & 33.10 & 42.51 & 32.38 & 47.56 \\
& SGA~\cite{lu2023set} & 92.71 & 93.02 & \textbf{100.0} & 99.96 & 37.01 & 46.62 & 41.43 & 51.72 \\
& DRA~\cite{gao2024boosting} & 95.45 & 95.36 & 99.91 & 99.98 & 47.69 & 57.06 & 52.46 & 62.51 \\
& TMM~\cite{wang2024transferable} & 68.10 & 72.30 & 97.87 & 97.60 & 54.63 & 63.52 & 58.87 & 65.71 \\
& SA-AET~\cite{jia2025semantic} & \textbf{98.73} & 98.41 & \textbf{100.0} & \textbf{100.0} & 56.03 & 63.61 & 59.55 & 67.13 \\
& GLEAM~\cite{liu2025gleam} & 98.17 & \textbf{98.46} & \textbf{100.0} & 99.90 & 66.11 & 66.90 & 66.97 & 67.71 \\
& \textbf{DeBias-Attack} & 93.43 & 94.60 & \textbf{100.0} & \textbf{100.0} & \textbf{68.10} & \textbf{72.65} & \textbf{69.35} & \textbf{76.47} \\
\midrule

\multirow{9}{*}{\textbf{CLIP$_{\rm ViT}$}}
& PGD~\cite{madry2017towards} & 3.32 & 6.19 & 4.80 & 8.55 & 69.64 & 84.32 & 13.45 & 17.22 \\
& BERT-Attack~\cite{li2020bert} & 9.83 & 23.10 & 11.59 & 24.70 & 28.77 & 38.79 & 30.63 & 37.20 \\
& Co-Attack~\cite{zhang2022towards} & 8.92 & 19.86 & 10.30 & 21.00 & 78.21 & 87.93 & 29.67 & 38.70 \\
& SGA~\cite{lu2023set} & 22.84 & 34.01 & 25.34 & 36.18 & \textbf{100.0} & \textbf{100.0} & 53.70 & 61.45 \\
& DRA~\cite{gao2024boosting} & 28.12 & 43.27 & 28.16 & 44.35 & \textbf{100.0} & \textbf{100.0} & 65.09 & 69.71 \\
& TMM~\cite{wang2024transferable} & 25.87 & 39.27 & 28.03 & 38.14 & \textbf{100.0} & \textbf{100.0} & 56.21 & 65.21 \\
& SA-AET~\cite{jia2025semantic} & 36.40 & 50.63 & 39.05 & 51.36 & \textbf{100.0} & \textbf{100.0} & 71.33 & 74.24 \\
& GLEAM~\cite{liu2025gleam} & 51.83 & 62.23 & 50.32 & 60.95 & 99.88 & 99.94 & 79.25 & 80.43 \\
& \textbf{DeBias-Attack} & \textbf{61.52} & \textbf{67.21} & \textbf{65.02} & \textbf{69.98} & \textbf{100.0} & \textbf{100.0} & \textbf{83.01} & \textbf{84.12} \\
\midrule

\multirow{9}{*}{\textbf{CLIP$_{\rm CNN}$}}
& PGD~\cite{madry2017towards} & 2.10 & 6.39 & 4.72 & 8.59 & 5.73 & 12.42 & 89.54 & 92.87 \\
& BERT-Attack~\cite{li2020bert} & 8.65 & 22.89 & 12.14 & 25.76 & 26.83 & 37.20 & 30.62 & 39.92 \\
& Co-Attack~\cite{zhang2022towards} & 10.24 & 23.85 & 12.79 & 26.27 & 27.53 & 40.92 & 95.66 & 96.21 \\
& SGA~\cite{lu2023set} & 15.43 & 28.22 & 18.44 & 33.32 & 38.84 & 51.13 & 99.86 & 99.88 \\
& DRA~\cite{gao2024boosting} & 19.31 & 34.84 & 21.35 & 37.51 & 48.82 & 58.84 & 99.84 & 99.89 \\
& TMM~\cite{wang2024transferable} & 17.43 & 30.14 & 19.46 & 35.24 & 42.12 & 52.37 & 97.74 & 97.34 \\
& SA-AET~\cite{jia2025semantic} & 24.10 & 38.01 & 27.10 & 42.04 & 54.32 & 64.00 & \textbf{100.0} & 99.95 \\
& GLEAM~\cite{liu2025gleam} & 36.93 & 48.41 & 36.12 & 52.06 & 62.89 & 69.03 & \textbf{100.0} & 99.81 \\
& \textbf{DeBias-Attack} & \textbf{50.16} & \textbf{59.35} & \textbf{59.11} & \textbf{64.62} & \textbf{70.92} & \textbf{75.52} & \textbf{100.0} & \textbf{100.0} \\
\bottomrule[0.3mm]
\end{tabular}}
\end{table*}

Table~\ref{tab:cross_model_flickr} reports R@1 ASR on Flickr30K. DeBias-Attack remains competitive on identical and closely related surrogate-target pairs, where ASR can already be saturated, and shows clearer advantages when the surrogate and target come from different model families. This trend appears in both transfer directions: from fusion-based VLP models to CLIP targets and from CLIP models back to fusion-based targets.
The improvement is balanced across TR and IR, suggesting that the corrected perturbation does not overfit a single retrieval direction. Since text retrieval and image retrieval query different sides of the aligned representation space, consistent gains in both directions indicate better transferability of cross-modal misalignment. This behavior is consistent with the motivation of DeBias-Attack, where reducing the component aligned with a weak-semantic reference gradient preserves attack strength on the surrogate while making the update less dependent on architecture-specific responses.

\begin{table*}[!t]
\caption{R@1 attack success rates (\%) on MSCOCO for image and text retrieval. ``Surrogate'' denotes the VLP model used to generate adversarial examples; higher values indicate stronger attacks, and bold values mark the best result for each surrogate-target retrieval setting (ties included).}
\label{tab:cross_model_mscoco}
\centering
\footnotesize
\renewcommand{\arraystretch}{1}
\setlength{\tabcolsep}{8pt}
\resizebox{0.95\textwidth}{!}{
\begin{tabular}{@{\extracolsep{\fill}} c|c|cc|cc|cc|cc}
\toprule[0.3mm]
& & \multicolumn{2}{c}{\textbf{ALBEF}} & \multicolumn{2}{c}{\textbf{TCL}} & \multicolumn{2}{c}{\textbf{CLIP$_{\rm ViT}$}} & \multicolumn{2}{c}{\textbf{CLIP$_{\rm CNN}$}} \\
\cmidrule{3-10}
\multirow{-2}{*}{\textbf{Surrogate}} & \multirow{-2}{*}{\textbf{Attack}} & TR R@1 & IR R@1 & TR R@1 & IR R@1 & TR R@1 & IR R@1 & TR R@1 & IR R@1 \\
\midrule

\multirow{9}{*}{\textbf{ALBEF}}
& PGD~\cite{madry2017towards} & 94.78 & 92.81 & 33.82 & 36.41 & 21.88 & 27.33 & 24.01 & 30.64 \\
& BERT-Attack~\cite{li2020bert} & 24.61 & 35.72 & 24.11 & 33.90 & 45.18 & 52.01 & 47.56 & 54.32 \\
& Co-Attack~\cite{zhang2022towards} & 95.30 & 97.42 & 64.89 & 73.04 & 55.62 & 62.01 & 56.41 & 66.12 \\
& SGA~\cite{lu2023set} & 99.92 & 99.97 & 87.09 & 88.53 & 63.41 & 69.88 & 63.77 & 70.49 \\
& DRA~\cite{gao2024boosting} & 99.86 & 99.91 & 88.53 & 89.82 & 69.02 & 75.59 & 68.31 & 75.36 \\
& TMM~\cite{wang2024transferable} & 96.79 & 97.73 & 70.19 & 74.02 & 68.37 & 75.34 & 70.97 & 76.88 \\
& SA-AET~\cite{jia2025semantic} & \textbf{100.0} & 99.96 & 96.15 & 95.67 & 76.81 & 80.02 & 76.44 & 80.31 \\
& GLEAM~\cite{liu2025gleam} & \textbf{100.0} & \textbf{100.0} & \textbf{100.0} & \textbf{99.67} & 81.31 & 78.53 & 77.59 & 76.78 \\
& \cellcolor{gray!25}DeBias-Attack & \textbf{100.0} & \textbf{100.0} & 95.21 & 94.90 & \textbf{83.90} & \textbf{86.33} & \textbf{82.35} & \textbf{86.12} \\
\midrule

\multirow{9}{*}{\textbf{TCL}}
& PGD~\cite{madry2017towards} & 40.42 & 44.57 & 98.21 & 98.67 & 21.66 & 26.75 & 24.14 & 32.01 \\
& BERT-Attack~\cite{li2020bert} & 35.08 & 45.21 & 38.31 & 48.76 & 51.37 & 58.49 & 52.40 & 61.11 \\
& Co-Attack~\cite{zhang2022towards} & 50.09 & 60.01 & 91.92 & 95.22 & 32.91 & 42.43 & 32.18 & 47.63 \\
& SGA~\cite{lu2023set} & 92.46 & 93.18 & \textbf{100.0} & \textbf{100.0} & 59.46 & 65.08 & 60.39 & 67.56 \\
& DRA~\cite{gao2024boosting} & 94.98 & 95.71 & \textbf{100.0} & \textbf{100.0} & 70.32 & 74.68 & 70.11 & 76.82 \\
& TMM~\cite{wang2024transferable} & 73.62 & 78.38 & 97.00 & 97.92 & 70.60 & 78.98 & 73.27 & 80.21 \\
& SA-AET~\cite{jia2025semantic} & 97.62 & 98.23 & \textbf{100.0} & 99.99 & 76.35 & 79.53 & 75.71 & 80.92 \\
& GLEAM~\cite{liu2025gleam} & \textbf{100.0} & \textbf{100.0} & \textbf{100.0} & \textbf{100.0} & 80.24 & 78.80 & 80.35 & 76.11 \\
& \cellcolor{gray!25}DeBias-Attack & 95.49 & 96.09 & \textbf{100.0} & \textbf{100.0} & \textbf{82.91} & \textbf{85.54} & \textbf{83.86} & \textbf{86.50} \\
\midrule

\multirow{9}{*}{\textbf{CLIP$_{\rm ViT}$}}
& PGD~\cite{madry2017towards} & 10.03 & 14.12 & 12.45 & 15.52 & 82.58 & 90.24 & 21.89 & 29.78 \\
& BERT-Attack~\cite{li2020bert} & 20.61 & 29.23 & 21.27 & 29.33 & 45.28 & 51.41 & 44.72 & 53.40 \\
& Co-Attack~\cite{zhang2022towards} & 26.01 & 36.95 & 28.52 & 38.09 & 88.52 & 96.39 & 47.22 & 58.12 \\
& SGA~\cite{lu2023set} & 43.48 & 51.77 & 44.29 & 50.86 & \textbf{100.0} & \textbf{100.0} & 70.33 & 75.24 \\
& DRA~\cite{gao2024boosting} & 52.93 & 61.12 & 51.66 & 61.78 & \textbf{100.0} & \textbf{100.0} & 80.71 & 83.98 \\
& TMM~\cite{wang2024transferable} & 45.58 & 56.83 & 46.51 & 53.67 & 97.10 & 98.21 & 70.27 & 75.12 \\
& SA-AET~\cite{jia2025semantic} & 57.42 & 67.11 & 57.30 & 65.42 & \textbf{100.0} & \textbf{100.0} & 84.12 & 86.97 \\
& GLEAM~\cite{liu2025gleam} & 63.87 & 71.15 & 59.30 & 67.75 & \textbf{100.0} & \textbf{100.0} & 86.32 & 88.92 \\
& \cellcolor{gray!25}DeBias-Attack & \textbf{75.77} & \textbf{77.28} & \textbf{76.35} & \textbf{77.03} & \textbf{100.0} & \textbf{100.0} & \textbf{89.82} & \textbf{91.68} \\
\midrule

\multirow{9}{*}{\textbf{CLIP$_{\rm CNN}$}}
& PGD~\cite{madry2017towards} &  8.04 & 12.48 & 11.63 & 15.41 & 13.48 & 20.44 & 92.49 & 94.59 \\
& BERT-Attack~\cite{li2020bert} & 23.12 & 34.41 & 24.32 & 29.27 & 51.05 & 57.18 & 54.29 & 61.98 \\
& Co-Attack~\cite{zhang2022towards} & 29.78 & 41.13 & 32.01 & 43.02 & 53.41 & 60.24 & 97.64 & 98.31 \\
& SGA~\cite{lu2023set} & 36.59 & 46.34 & 38.47 & 48.71 & 62.07 & 67.59 & 99.89 & 99.97 \\
& DRA~\cite{gao2024boosting} & 41.40 & 52.25 & 43.62 & 54.15 & 70.43 & 74.14 & 99.80 & 99.92 \\
& TMM~\cite{wang2024transferable} & 38.67 & 54.21 & 36.56 & 55.83 & 64.51 & 68.12 & 97.27 & 97.78 \\
& SA-AET~\cite{jia2025semantic} & 44.01 & 55.42 & 47.01 & 57.39 & 73.67 & 76.90 & \textbf{100.0} & 99.92 \\
& GLEAM~\cite{liu2025gleam} & 49.53 & 57.17 & 52.51 & 60.60 & 75.98 & 81.12 & \textbf{100.0} & \textbf{100.0} \\
& \cellcolor{gray!25}DeBias-Attack & \textbf{66.72} & \textbf{69.89} & \textbf{69.21} & \textbf{71.50} & \textbf{85.23} & \textbf{87.75} & \textbf{100.0} & \textbf{100.0} \\
\bottomrule[0.3mm]
\end{tabular}}
\end{table*}

Table~\ref{tab:cross_model_mscoco} further reports MSCOCO results. The same pattern appears on this dataset, where DeBias-Attack remains stable in saturated related-model settings and shows clearer advantages under heterogeneous transfer, especially between fusion-based VLP models and CLIP-style models. Since Flickr30K and MSCOCO differ in image diversity and caption distribution, this consistency indicates that the debiased update is not tied to a particular retrieval benchmark.

\subsection{Cross-Task Attack Performance Evaluation}

\begin{table}[!t]
\caption{Performance comparison on the ITR $\rightarrow$ VG task. Lower values ($\downarrow$) indicate better attack performance, and bold values mark the best result in each column.}
\centering
\small
\renewcommand\arraystretch{1.1}
\setlength{\tabcolsep}{6pt}
\resizebox{0.85\linewidth}{!}{
\begin{tabular}{c|ccc}
\toprule
\textbf{Attack} & \textbf{Val $\downarrow$} & \textbf{TestA $\downarrow$} & \textbf{TestB $\downarrow$} \\
\midrule
Original Input & 58.46 & 65.89 & 46.25 \\
SGA~\cite{lu2023set} & 50.45 & 57.71 & 41.34 \\
DRA~\cite{gao2024boosting} & 49.02 & 56.23 & 42.21 \\
SA-AET~\cite{jia2025semantic} & 46.03 & 53.13 & 37.12 \\
GLEAM~\cite{liu2025gleam} & 45.51 & 54.34 & 36.84 \\
DeBias-Attack    & \textbf{44.31} & \textbf{50.52} & \textbf{36.14} \\
\bottomrule
\end{tabular}}
\label{tab:cross_task_itr_vg}
\end{table}

\begin{table}[!t]
\caption{Performance comparison on the ITR $\rightarrow$ IC task. Lower values ($\downarrow$) indicate better attack performance, and bold values mark the best result in each column.}
\centering
\small
\renewcommand\arraystretch{1.1}
\setlength{\tabcolsep}{6pt}
\resizebox{0.95\linewidth}{!}{
\begin{tabular}{c|ccccc}
\toprule
 & \textbf{B@4 $\downarrow$} & \textbf{METEOR $\downarrow$} & \textbf{ROUGE-L $\downarrow$} & \textbf{CIDEr $\downarrow$} & \textbf{SPICE $\downarrow$} \\
\midrule
Original Input & 39.7 & 31.0 & 60.0 & 133.3 & 23.8 \\
SGA~\cite{lu2023set} & 29.1 & 24.3 & 51.5 & 91.2 & 17.8 \\
DRA~\cite{gao2024boosting} & 26.6 & 24.6 & 48.6 & 86.9 & 16.9 \\
SA-AET~\cite{jia2025semantic} & 20.8 & 20.5 & 44.7 & 65.9 & 13.4 \\
GLEAM~\cite{liu2025gleam} & 19.8 & 19.3 & 42.8 & 59.7 & 12.4 \\
DeBias-Attack & \textbf{18.1} & \textbf{18.7} & \textbf{42.8} & \textbf{56.3} & \textbf{11.9} \\
\bottomrule
\end{tabular}}
\label{tab:cross_task_itr_ic}
\end{table}

We next test whether adversarial examples generated for ITR transfer to VG and IC without additional optimization. For this cross-task evaluation, we reproduce and compare the strongest baselines from Table~\ref{tab:cross_model_flickr}, including SGA, DRA, SA-AET, and GLEAM. Table~\ref{tab:cross_task_itr_vg} shows that DeBias-Attack reduces validation accuracy from 58.46 to 44.31 and TestA accuracy from 65.89 to 50.52, outperforming GLEAM by 1.91 points on average. Table~\ref{tab:cross_task_itr_ic} shows drops across captioning metrics, reducing CIDEr from 133.3 to 56.3 and SPICE from 23.8 to 11.9, while improving over GLEAM on B@4, METEOR, CIDEr, and SPICE. Retrieval-oriented adversarial examples transfer to localization and generation, not only to another ranking metric.
VG requires fine-grained region-text correspondence, whereas IC requires generating fluent descriptions from visual content. Degradation on both tasks suggests that the adversarial image-text pair disrupts semantic alignment at a level that remains relevant after the downstream objective changes. This is a stronger transfer test than cross-model retrieval alone because the victim output space and evaluation metrics are no longer the same as the attack-generation objective.

\subsection{Ablation Study}

\begin{table*}[!t]
\caption{Input ablation for the reference adversarial gradient on Flickr30K with ALBEF as the surrogate. TR/IR report R@1 ASR (\%); higher is better, bold marks column-wise best values (ties included), and parentheses show drops from the best.}
\label{tab:ablation_input}
\centering
\scriptsize
\renewcommand{\arraystretch}{1.08}
\setlength{\tabcolsep}{3.8pt}
\begin{tabular*}{\textwidth}{@{\extracolsep{\fill}}lcccccccc}
\toprule
\multirow{2}{*}{Reference input} &
\multicolumn{2}{c}{ALBEF} &
\multicolumn{2}{c}{TCL} &
\multicolumn{2}{c}{CLIP$_{\rm ViT}$} &
\multicolumn{2}{c}{CLIP$_{\rm CNN}$} \\
\cmidrule(lr){2-3} \cmidrule(lr){4-5} \cmidrule(lr){6-7} \cmidrule(lr){8-9}
& TR R@1 & IR R@1 & TR R@1 & IR R@1 & TR R@1 & IR R@1 & TR R@1 & IR R@1 \\
\midrule
Mean + Low-var. noise
& \textbf{100.00} & \textbf{100.00}
& \textbf{91.89} & \textbf{92.33}
& \textbf{70.06} & \textbf{73.26}
& \textbf{70.88} & \textbf{74.96} \\
Low-var. noise
& \textbf{100.00} & \textbf{100.00}
& 91.25 ($\downarrow$0.64) & 91.24 ($\downarrow$1.09)
& 68.71 ($\downarrow$1.35) & 72.71 ($\downarrow$0.55)
& 70.26 ($\downarrow$0.62) & 74.51 ($\downarrow$0.45) \\
Mean + High-var. noise
& \textbf{100.00} & \textbf{100.00}
& 90.20 ($\downarrow$1.69) & 90.71 ($\downarrow$1.62)
& 67.36 ($\downarrow$2.70) & 71.26 ($\downarrow$2.00)
& 67.69 ($\downarrow$3.19) & 72.73 ($\downarrow$2.23) \\
High-var. noise
& \textbf{100.00} & \textbf{100.00}
& 88.94 ($\downarrow$2.95) & 90.55 ($\downarrow$1.78)
& 66.01 ($\downarrow$4.05) & 71.36 ($\downarrow$1.90)
& 69.22 ($\downarrow$1.66) & 72.59 ($\downarrow$2.37) \\
Mean
& \textbf{100.00} & \textbf{100.00}
& 88.83 ($\downarrow$3.06) & 90.45 ($\downarrow$1.88)
& 65.89 ($\downarrow$4.17) & 71.42 ($\downarrow$1.84)
& 69.35 ($\downarrow$1.53) & 72.45 ($\downarrow$2.51) \\
\bottomrule
\end{tabular*}
\end{table*}

We ablate the input used to optimize the reference perturbation branch with ALBEF on Flickr30K. Table~\ref{tab:ablation_input} compares the dataset mean image, a uniform 0.5 image with Gaussian noise, and the dataset mean image with Gaussian noise. The 0.5-prior input removes image-specific semantics while keeping the reference input within the normalized pixel range, whereas the dataset mean preserves coarse dataset-level statistics. Low-variance noise improves heterogeneous transfer, while high variance gives limited gains, suggesting that excessive randomness weakens the reference signal. The dataset mean with low-variance noise performs best on every non-source target and retrieval direction.
The reference input should be weak-semantic. An artificial prior can reduce image semantics, but it may produce gradients detached from natural-image statistics. Mild noise around the dataset mean keeps the reference branch close to the data distribution while suppressing instance-level semantics, which makes the reference gradient better suited for estimating surrogate-specific bias. The ablation therefore supports both parts of the reference-branch design, showing that weak semantics are needed to expose surrogate-preferred update directions and that dataset-level statistics help keep this estimate relevant to natural images.

\begin{table*}[!t]
\caption{R@1 attack success rates (\%) on Flickr30K with ablations on the correction strength $\beta$ using ALBEF as the surrogate model. TR and IR denote text retrieval and image retrieval, respectively. Higher values indicate stronger attacks, and bold values mark the best result in each column.}
\label{tab:ablation_beta}
\centering
\footnotesize
\renewcommand{\arraystretch}{1}
\setlength{\tabcolsep}{8pt}
\begin{tabular*}{0.95\textwidth}{@{\extracolsep{\fill}} c|cc|cc|cc|cc}
\toprule[0.3mm]
& \multicolumn{2}{c}{\textbf{ALBEF}} & \multicolumn{2}{c}{\textbf{TCL}} & \multicolumn{2}{c}{\textbf{CLIP$_{\rm ViT}$}} & \multicolumn{2}{c}{\textbf{CLIP$_{\rm CNN}$}} \\
\cmidrule{2-9}
\multirow{-2}{*}{\textbf{$\beta$}} & TR R@1 & IR R@1 & TR R@1 & IR R@1 & TR R@1 & IR R@1 & TR R@1 & IR R@1 \\
\midrule
0 & \textbf{100.0} & \textbf{100.0} & \textbf{93.26} & \textbf{93.33} & 56.93 & 65.85 & 59.26 & 66.69 \\
0.5 & \textbf{100.0} & \textbf{100.0} & 92.25 & 91.43 & 61.10 & 67.91 & 64.62 & 70.57 \\
0.7 & \textbf{100.0} & \textbf{100.0} & 91.21 & 90.62 & 63.93 & 69.20 & 67.05 & 72.01 \\
0.8 & \textbf{100.0} & \textbf{100.0} & 91.30 & 91.69 & 66.47 & 71.82 & 67.13 & 72.62 \\
\textbf{0.9} & \textbf{100.0} & \textbf{100.0} & 91.89 & 92.33 & \textbf{70.06} & \textbf{73.26} & \textbf{70.88} & \textbf{74.96} \\
1.0 & \textbf{100.0} & \textbf{100.0} & 89.46 & 91.31 & 63.93 & 69.46 & 66.92 & 71.70 \\
1.3 & \textbf{100.0} & \textbf{100.0} & 89.15 & 91.19 & 63.56 & 69.52 & 66.82 & 71.60 \\
1.5 & \textbf{100.0} & \textbf{100.0} & 88.72 & 90.57 & 61.60 & 67.26 & 64.50 & 70.67 \\
\bottomrule[0.3mm]
\end{tabular*}
\end{table*}

We also ablate the correction strength $\beta$. Table~\ref{tab:ablation_beta} shows that $\beta=0$ weakens transfer on CLIP-based targets, while increasing $\beta$ improves heterogeneous transfer by suppressing the aligned reference-gradient component. The best performance is obtained at $\beta=0.9$, which achieves the highest ASR on both CLIP$_{\rm ViT}$ and CLIP$_{\rm CNN}$ while preserving white-box attack performance on ALBEF. Larger values reduce ASR, suggesting that strong correction can also suppress useful adversarial components.
$\beta$ controls a trade-off between debiasing and attack strength. Moderate correction removes the component most likely to be surrogate-specific, but excessive correction can distort the main adversarial direction and remove useful perturbation components. The empirical optimum at $\beta=0.9$ matches this trade-off.

\subsection{Transfer to Multimodal Large Language Models}

\begin{table}[!t]
\caption{Performance comparison on MLLM attack transfer. Higher ASR and mASR ($\uparrow$) and lower Sim. ($\downarrow$) indicate stronger attacks, and bold values mark the best result in each column.}
\centering
\small
\renewcommand\arraystretch{1.1}
\setlength{\tabcolsep}{6pt}
\resizebox{0.90\linewidth}{!}{
\begin{tabular}{l l c c c}
\toprule
MLLM & Method & ASR$\uparrow$ & Sim.$\downarrow$ & mASR$\uparrow$ \\
\midrule
\multirow{2}{*}{BLIP-2}
& SA-AET        & 3.70  & 0.3108 & 44.74 \\
& DeBias-Attack & \textbf{10.50} & \textbf{0.2973} & \textbf{51.29} \\
\midrule
\multirow{2}{*}{Qwen2-VL}
& SA-AET        & 19.90 & 0.3151 & 42.69 \\
& DeBias-Attack & \textbf{30.80} & \textbf{0.3017} & \textbf{49.08} \\
\midrule
\multirow{2}{*}{Claude-3.5}
& SA-AET        & 52.71 & 0.3123 & 44.05 \\
& DeBias-Attack & \textbf{67.71} & \textbf{0.2981} & \textbf{50.95} \\
\midrule
\multirow{2}{*}{GPT-4o}
& SA-AET        & 58.70 & 0.3390 & 31.18 \\
& DeBias-Attack & \textbf{67.46} & \textbf{0.3299} & \textbf{35.53} \\
\midrule
\multirow{2}{*}{GPT-4o-mini}
& SA-AET        & 55.36 & 0.3328 & 34.05 \\
& DeBias-Attack & \textbf{62.37} & \textbf{0.3264} & \textbf{37.11} \\
\bottomrule
\end{tabular}}
\label{tab:mllm_quantitative}
\end{table}

We evaluate black-box captioning attacks on BLIP-2, Qwen2-VL, Claude-3.5, GPT-4o, and GPT-4o-mini. Adversarial examples are generated on ALBEF, and each model is prompted to ``describe this picture within 50 words.'' Because MLLM outputs are free-form, lexical overlap alone may miss semantic failures. We therefore measure semantic alignment with CLIP image-text cosine similarity and count success when adversarial similarity falls below the corresponding mean similarity on original inputs. Additional qualitative examples, including Qwen3-VL results, are provided in Appendix~\ref{app:qualitative}.

Table~\ref{tab:mllm_quantitative} shows that DeBias-Attack achieves higher ASR, lower similarity, and higher mASR than SA-AET on all five MLLMs. It improves ASR from 52.71\% to 67.71\% on Claude-3.5 and from 58.70\% to 67.46\% on GPT-4o. Since mASR averages success rates over reference similarity levels from 0.20 to 0.40 with a step size of 0.005, the advantage is not tied to a single threshold. Similar improvements on BLIP-2, Qwen2-VL, and GPT-4o-mini show that the perturbations remain effective beyond standard VLP evaluation models.
These MLLM targets differ from the surrogate VLP model in architecture, training data, and generation behavior. The reduction in semantic similarity suggests that debiased perturbations transfer beyond retrieval models and can influence open-ended multimodal generation. This result is important from a security perspective because the attack is optimized only against a retrieval-style VLP surrogate but still affects instruction-following multimodal systems.

\begin{table}[!t]
\caption{Sample-level comparison with SA-AET on MLLM attack transfer. Wins and losses are counted per sample; higher win rate ($\uparrow$) indicates better attack performance, and bold values mark the reported win rates.}
\centering
\small
\renewcommand\arraystretch{1.1}
\setlength{\tabcolsep}{6pt}
\resizebox{0.80\linewidth}{!}{
\begin{tabular}{lccc}
\toprule
MLLM & Wins & Losses & Win Rate$\uparrow$ \\
\midrule
BLIP-2       & 603 & 252 & \textbf{70.53} \\
Qwen2-VL     & 634 & 354 & \textbf{64.17} \\
Claude-3.5   & 322 & 158 & \textbf{67.08} \\
GPT-4o       & 576 & 347 & \textbf{62.41} \\
GPT-4o-mini  & 549 & 420 & \textbf{56.66} \\
\bottomrule
\end{tabular}}
\label{tab:mllm_pairwise_win}
\end{table}

We also conduct a sample-level comparison with SA-AET on Flickr30K. For each image, DeBias-Attack is counted as a win if it produces a lower CLIP image-text cosine similarity than SA-AET; ties are omitted. Table~\ref{tab:mllm_pairwise_win} records more wins than losses on all evaluated MLLMs, including 603 versus 252 on BLIP-2, 634 versus 354 on Qwen2-VL, and 576 versus 347 on GPT-4o. The advantage is distributed broadly across samples, indicating a consistent sample-level gain. This pairwise view reduces dependence on a single success threshold and confirms that the improvement is visible at the individual-example level.

\section{Conclusion}
\label{sec:conclusion}

We introduced DeBias-Attack, a transfer-based adversarial attack framework for VLP models. DeBias-Attack addresses surrogate-specific bias through bias-guided gradient correction during perturbation optimization. It learns a main perturbation on the original image and a reference perturbation on randomized weak-semantic images. The reference perturbation is separated from the main perturbation and evaluated on the original image to provide a gradient that reflects surrogate-specific bias. The debiased perturbation is updated by combining its own adversarial gradient with the corrected direction derived from the main and reference gradients, while the textual perturbation is refined with the updated debiased adversarial image. The generated adversarial image-text pairs preserve attack strength while reducing dependence on surrogate-specific bias.
Experiments show that DeBias-Attack improves black-box transferability across VLP models, downstream tasks, and real-world multimodal large language models, with clear gains under heterogeneous model transfer. Correcting surrogate-specific bias during optimization improves transfer-based VLP attacks. Future work will extend this direction to broader multimodal scenarios and investigate defenses that can identify or reduce transferable cross-modal semantic misalignment.

\section*{Acknowledgments}
\label{sec:acknowledgments}

This work is supported by the National Natural Science Foundation of China under Grants No.62472092, No.62172089, and No.62106045; the Natural Science Foundation of Jiangsu Province under Grant No.BK20241751; Jiangsu Provincial Key Laboratory of Computer Networking Technology; Jiangsu Provincial Key Laboratory of Network and Information Security under Grant No.BM2003201; Key Laboratory of Computer Network and Information Integration of Ministry of Education of China under Grant No.93K-9; Nanjing Purple Mountain Laboratories; and Fintech and Big Data Laboratory of Southeast University. We thank the Big Data Computing Center of Southeast University for computational support.

\bibliographystyle{IEEEtran}
\bibliography{references}

\begin{IEEEbiography}[{\includegraphics[width=1in,height=1.25in,clip,keepaspectratio]{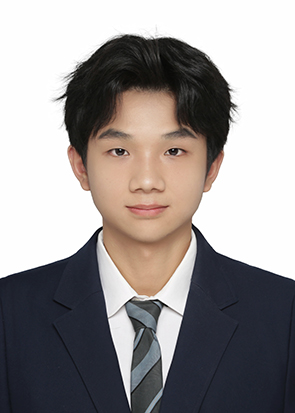}}]{Lijia Yu}
received the B.S. degree in computer science and technology from Nanchang University, Nanchang, China, in 2023. He is currently pursuing the Ph.D. degree with the School of Cyber Science and Engineering, Southeast University, Nanjing, China, under the supervision of Prof. Jiuxin Cao. His research interests include the security and robustness of multimodal models, adversarial attacks, robustness evaluation, and trustworthy multimodal learning.
\end{IEEEbiography}

\begin{IEEEbiography}[{\includegraphics[width=1in,height=1.25in,clip,keepaspectratio]{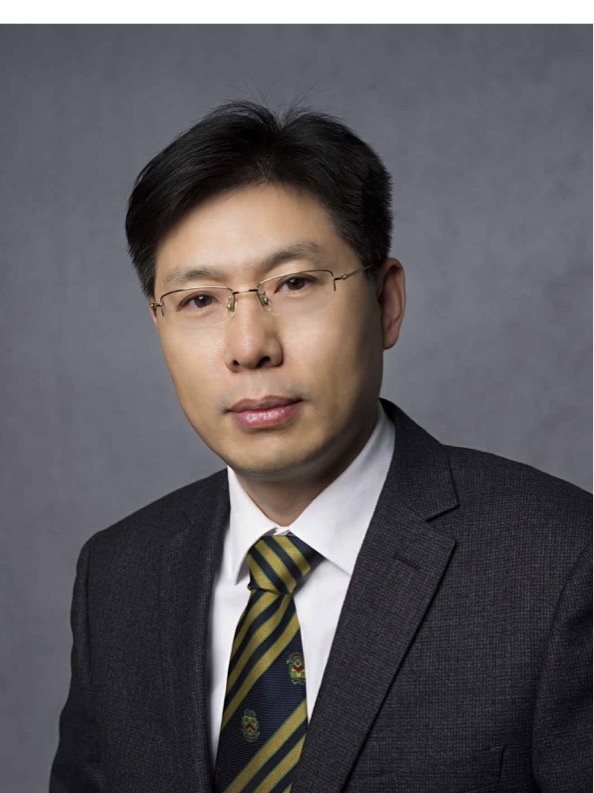}}]{Jiuxin Cao}
received the Ph.D. degree in computer science and technology from Xi'an Jiaotong University, Xi'an, China, in 2003. He is currently a Full Professor with the School of Cyber Science and Engineering, Southeast University, China. He is the Leader of the Jiangsu Provincial Key Laboratory of Computer Network Technology and the principal investigator of several national projects in China. He has published more than 100 papers in journals and conferences, including IJCAI, ACM MM, WWW, TIP, and IEEE TCSS. His main research interests include AI security, computational society, and cross-modality media fusion.
\end{IEEEbiography}

\begin{IEEEbiography}[{\includegraphics[width=1in,height=1.25in,clip,keepaspectratio]{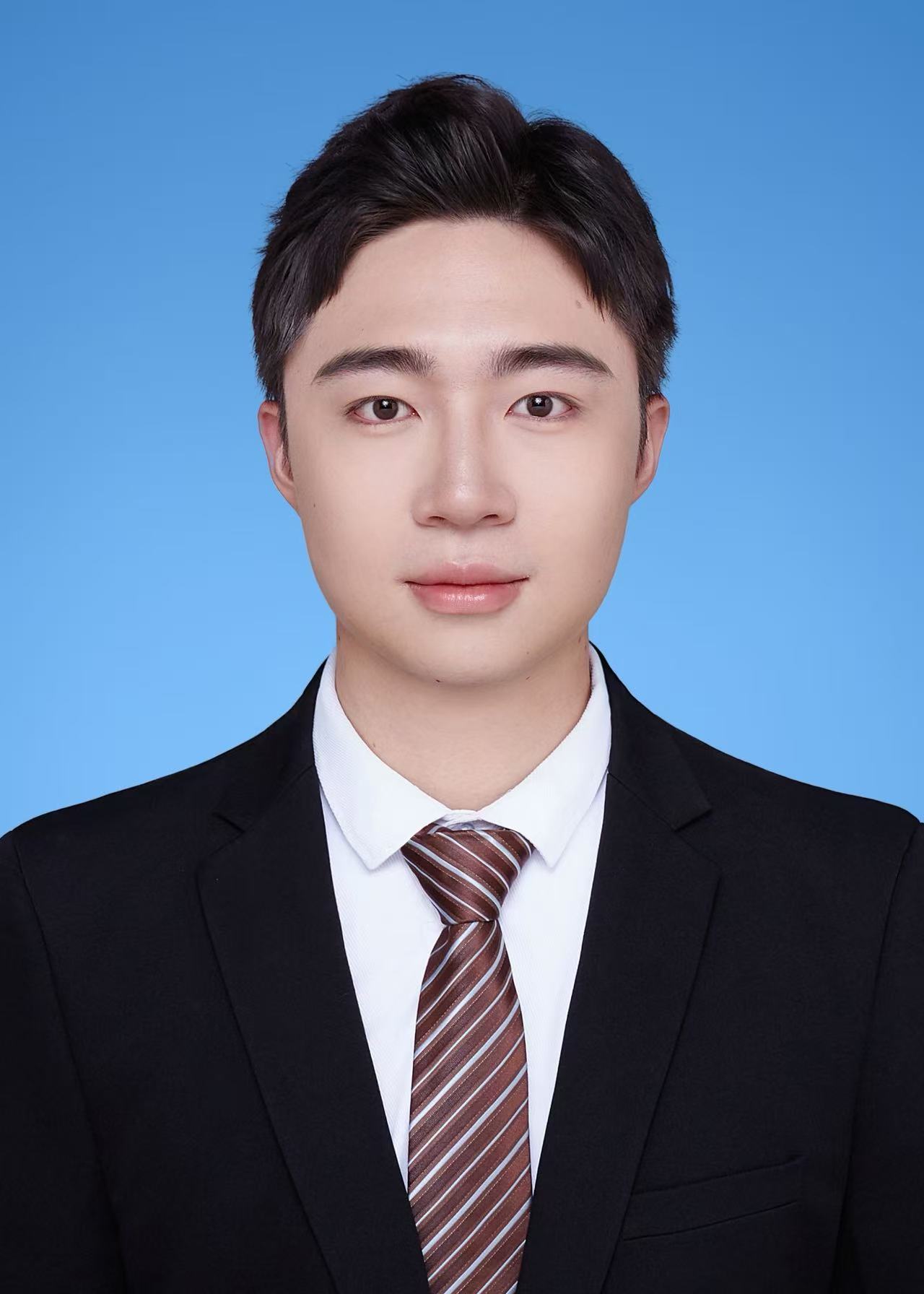}}]{Yuchen Qiang}
received the B.S. degree in Internet of Things from Jiangsu University, Zhenjiang, China, in 2023. He is currently pursuing the Ph.D. degree with the School of Cyber Science and Engineering, Southeast University, Nanjing, China, under the supervision of Prof. Jiuxin Cao. His research interests include anomaly detection, vision representation learning, and industrial applications. He works on algorithms and systems for anomaly detection in industrial environments and vision-based real-time monitoring.
\end{IEEEbiography}

\begin{IEEEbiography}[{\includegraphics[width=1in,height=1.25in,clip,keepaspectratio]{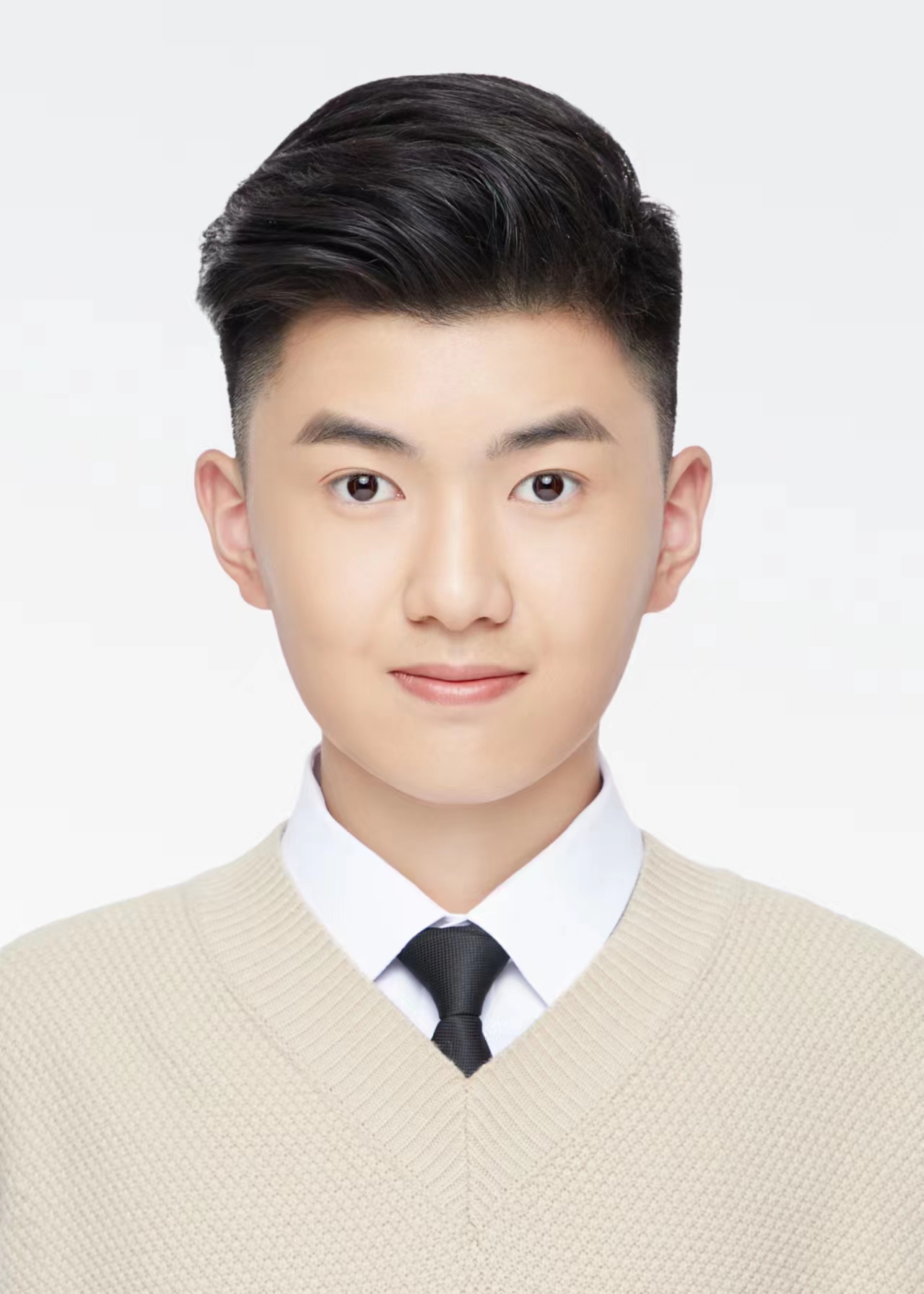}}]{Changhao Chen}
received the B.S. degree in cyber science and engineering from Southeast University, Nanjing, China, in 2025. He is currently pursuing the M.S. degree in cyber science and engineering with the School of Cyber Science and Engineering, Southeast University, under the supervision of Prof. Jiuxin Cao. His work focuses on multimodal model security, including adversarial attacks and defenses for multimodal learning systems. His main research interests include multimodal learning, AI security, and the adversarial robustness of visual and multimodal models.
\end{IEEEbiography}

\begin{IEEEbiography}[{\includegraphics[width=1in,height=1.25in,clip,keepaspectratio]{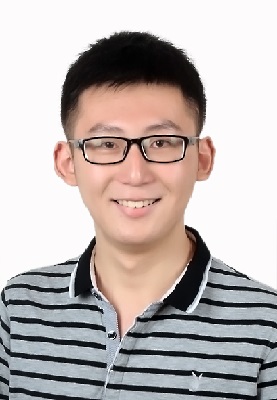}}]{Yifei Huang}
received the B.S. degree in cyber science and engineering from Southeast University, Nanjing, China, in 2022. He is currently working toward the Ph.D. degree in cyber science and engineering with the School of Cyber Science and Engineering, Southeast University, Nanjing, China, under the supervision of Prof. Jiuxin Cao. His research interests include natural language processing, large language model-generated text augmentation, large language model-generated text detection and provenance tracing, and large language model text watermarking.
\end{IEEEbiography}

\begin{IEEEbiography}[{\includegraphics[width=1in,height=1.25in,clip,keepaspectratio]{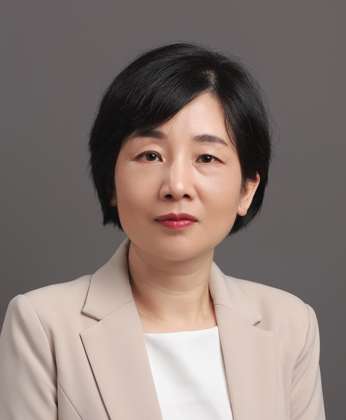}}]{Bo Liu}
received the Ph.D. degree from Southeast University, Nanjing, China. She is currently a Full Professor with the School of Computer Science and Engineering, Southeast University. She is the principal investigator of several national projects in China. She has published over 80 papers in major journals and international conferences, such as ACL, WWW, KDD, IEEE/ACM Transactions on Networking (TON), and IEEE Transactions on Neural Networks and Learning Systems (TNNLS). Her current research interests include network anomaly detection, network behavior analysis, user behavior analysis in social networks, trustworthiness of large language models, and social influence.
\end{IEEEbiography}

\vfill

\clearpage
\clearpage

\twocolumn[{
\begin{center}
{\LARGE Appendix}
\end{center}
\vspace{0.5em}
}]
\normalfont\normalsize

\setcounter{section}{0}
\makeatletter
\@addtoreset{equation}{section}
\@addtoreset{figure}{section}
\@addtoreset{table}{section}
\makeatother
\renewcommand{\thesection}{\Alph{section}}
\renewcommand{\theequation}{\thesection.\arabic{equation}}
\renewcommand{\thefigure}{\thesection.\arabic{figure}}
\renewcommand{\thetable}{\thesection.\arabic{table}}

\section{Theoretical Analysis of DeBias-Attack}
\label{app:theory}

We analyze why bias-guided gradient correction can reduce surrogate-specific components and improve black-box transferability. The analysis focuses on image-side optimization and omits the text update. It states the corrected update rule and derives a condition under which this correction improves the loss of a held-out target model under a first-order approximation.

\textbf{Proposition 1 (Reference-gradient correction).}
At iteration $t$, let $l_s(\cdot,\cdot)$ denote the projected misalignment score of the surrogate model. The main adversarial gradient and the reference adversarial gradient are
\begin{equation}
\begin{aligned}
    \nabla_m^{(t)}
    &=
    \nabla_{\boldsymbol{\delta}_{m}^{(t)}}
    l_s\!\left(
    \mathbf{x}_{I}
    +
    \boldsymbol{\delta}_{m}^{(t)},
    \tilde{\mathbf{x}}_{T}^{(t)}
    \right), \\
    \nabla_r^{(t)}
    &=
    \nabla_{\boldsymbol{\delta}_{r}^{(t)}}
    l_s\!\left(
    \mathbf{x}_{I}
    +
    \boldsymbol{\delta}_{r}^{(t)},
    \tilde{\mathbf{x}}_{T}^{(t)}
    \right).
\end{aligned}
\label{eq:app_main_reference_grad}
\end{equation}
The positive projection coefficient and the corrected direction are defined as
\begin{equation}
    c^{(t)}
    =
    \frac{
    \max\!\left(
    \left\langle
    \nabla_m^{(t)},
    \nabla_r^{(t)}
    \right\rangle,
    0
    \right)
    }{
    \left\|\nabla_r^{(t)}\right\|_2^2
    +
    \epsilon_{\mathrm{num}}
    },
    \label{eq:app_projection_coeff}
\end{equation}
\begin{equation}
    \mathbf{d}_{\mathrm{corr}}^{(t)}
    =
    \nabla_m^{(t)}
    -
    \beta c^{(t)}\nabla_r^{(t)},
    \label{eq:app_corrected_direction}
\end{equation}
where $\beta>0$ is the correction strength and $\epsilon_{\mathrm{num}}$ is a small numerical constant. Then the corrected direction satisfies
\begin{equation}
\begin{aligned}
    \left\langle
    \mathbf{d}_{\mathrm{corr}}^{(t)},
    \nabla_r^{(t)}
    \right\rangle
    &=
    \left\langle
    \nabla_m^{(t)},
    \nabla_r^{(t)}
    \right\rangle
    -
    \beta c^{(t)}
    \left\|\nabla_r^{(t)}\right\|_2^2 .
\end{aligned}
\label{eq:app_alignment_reduction}
\end{equation}
When $\left\langle\nabla_m^{(t)},\nabla_r^{(t)}\right\rangle>0$, the overlap with the reference adversarial gradient is reduced by a non-negative amount. When $\left\langle\nabla_m^{(t)},\nabla_r^{(t)}\right\rangle\leq0$, $c^{(t)}=0$ and no correction is applied.

\textit{Proof.}
Substituting Eq.~\eqref{eq:app_projection_coeff} into Eq.~\eqref{eq:app_corrected_direction}, we have
\begin{equation}
\begin{aligned}
    \left\langle
    \mathbf{d}_{\mathrm{corr}}^{(t)},
    \nabla_r^{(t)}
    \right\rangle
    &=
    \left\langle
    \nabla_m^{(t)},
    \nabla_r^{(t)}
    \right\rangle
    -
    \beta c^{(t)}
    \left\langle
    \nabla_r^{(t)},
    \nabla_r^{(t)}
    \right\rangle \\
    &=
    \left\langle
    \nabla_m^{(t)},
    \nabla_r^{(t)}
    \right\rangle
    -
    \beta c^{(t)}
    \left\|\nabla_r^{(t)}\right\|_2^2 .
\end{aligned}
\label{eq:app_correction_proof}
\end{equation}
This proves Eq.~\eqref{eq:app_alignment_reduction}. If the main and reference gradients are positively aligned, $c^{(t)}>0$ and the second term in Eq.~\eqref{eq:app_alignment_reduction} is non-negative. Otherwise, the maximum operator in Eq.~\eqref{eq:app_projection_coeff} gives $c^{(t)}=0$. This completes the proof.

\textbf{Proposition 2 (Projected-sign implementation).}
Let the continuous raw and corrected update directions be
\begin{equation}
\begin{aligned}
    \mathbf{u}_{\mathrm{raw}}^{(t)}
    &=
    \nabla_{\boldsymbol{\delta}_{I}^{(t)}}
    l_s\!\left(
    \mathbf{x}_{I}
    +
    \boldsymbol{\delta}_{I}^{(t)},
    \tilde{\mathbf{x}}_{T}^{(t)}
    \right)
    +
    \nabla_m^{(t)}, \\
    \mathbf{u}^{(t)}
    &=
    \nabla_{\boldsymbol{\delta}_{I}^{(t)}}
    l_s\!\left(
    \mathbf{x}_{I}
    +
    \boldsymbol{\delta}_{I}^{(t)},
    \tilde{\mathbf{x}}_{T}^{(t)}
    \right)
    +
    \mathbf{d}_{\mathrm{corr}}^{(t)} .
\end{aligned}
\label{eq:app_raw_and_corrected_update}
\end{equation}
Define the signed directions as
\begin{equation}
    \mathbf{s}_{\mathrm{raw}}^{(t)}
    =
    \mathrm{sign}\!\left(\mathbf{u}_{\mathrm{raw}}^{(t)}\right),
    \quad
    \mathbf{s}_{\mathrm{corr}}^{(t)}
    =
    \mathrm{sign}\!\left(\mathbf{u}^{(t)}\right),
    \label{eq:app_signed_directions}
\end{equation}
and the actual projected displacement induced by a signed direction $\mathbf{s}$ as
\begin{equation}
    \mathbf{q}(\mathbf{s})
    =
    \Pi_{\varepsilon}\!\left(
    \boldsymbol{\delta}_{I}^{(t)}
    +
    \alpha\mathbf{s}
    \right)
    -
    \boldsymbol{\delta}_{I}^{(t)} .
    \label{eq:app_projected_step}
\end{equation}
Away from zero-sign boundary points and non-informative saturated coordinates, the implemented projected-sign correction does not increase the overlap with the reference gradient:
\begin{equation}
    \left\langle
    \mathbf{q}\!\left(\mathbf{s}_{\mathrm{corr}}^{(t)}\right),
    \nabla_r^{(t)}
    \right\rangle
    \leq
    \left\langle
    \mathbf{q}\!\left(\mathbf{s}_{\mathrm{raw}}^{(t)}\right),
    \nabla_r^{(t)}
    \right\rangle .
    \label{eq:app_projected_alignment_reduction}
\end{equation}

\textit{Proof.}
By Eq.~\eqref{eq:app_corrected_direction}, we have
\begin{equation}
    \mathbf{u}^{(t)}
    =
    \mathbf{u}_{\mathrm{raw}}^{(t)}
    -
    \beta c^{(t)}\nabla_r^{(t)} .
    \label{eq:app_update_relation}
\end{equation}
For each coordinate $j$, the corrected update shifts $u_{\mathrm{raw},j}^{(t)}$ in the direction opposite to $\nabla_{r,j}^{(t)}$. Hence
\begin{equation}
    \mathrm{sign}\!\left(u_j^{(t)}\right)
    \nabla_{r,j}^{(t)}
    \leq
    \mathrm{sign}\!\left(u_{\mathrm{raw},j}^{(t)}\right)
    \nabla_{r,j}^{(t)} .
    \label{eq:app_coordinate_sign_reduction}
\end{equation}
Summing Eq.~\eqref{eq:app_coordinate_sign_reduction} over all coordinates gives
\begin{equation}
    \left\langle
    \mathbf{s}_{\mathrm{corr}}^{(t)},
    \nabla_r^{(t)}
    \right\rangle
    \leq
    \left\langle
    \mathbf{s}_{\mathrm{raw}}^{(t)},
    \nabla_r^{(t)}
    \right\rangle .
    \label{eq:app_signed_alignment_reduction}
\end{equation}
The projection $\Pi_{\varepsilon}$ clips each coordinate separately. It can reduce the magnitude of a signed step but does not reverse its sign. After removing coordinates that are already saturated and cannot move, Eq.~\eqref{eq:app_signed_alignment_reduction} implies Eq.~\eqref{eq:app_projected_alignment_reduction}. This completes the proof.

\textbf{Theorem 1 (Target-loss improvement condition).}
Let $l_{\tau}(\cdot,\cdot)$ denote the projected misalignment score of a held-out target model $\tau$, and let
\begin{equation}
    \nabla_{\tau}^{(t)}
    =
    \nabla_{\mathbf{x}}
    l_{\tau}\!\left(
    \mathbf{x}_{I}
    +
    \boldsymbol{\delta}_{I}^{(t)},
    \tilde{\mathbf{x}}_{T}^{(t)}
    \right)
    \label{eq:app_target_grad}
\end{equation}
be the local target gradient. Under a first-order approximation, the corrected update improves the target-model misalignment score over the uncorrected update if
\begin{equation}
    \alpha>0,
    \quad
    \beta>0,
    \quad
    c^{(t)}>0,
    \quad
    \left\langle
    \nabla_{\tau}^{(t)},
    \nabla_r^{(t)}
    \right\rangle
    <0 .
    \label{eq:app_sufficient_condition}
\end{equation}

\textit{Proof.}
By the first-order Taylor approximation, one image-side update changes the target-model loss by
\begin{equation}
    \Delta l_{\tau}^{(t)}
    \approx
    \alpha
    \left\langle
    \nabla_{\tau}^{(t)},
    \mathbf{u}^{(t)}
    \right\rangle .
    \label{eq:app_target_change}
\end{equation}
Meanwhile, the corresponding raw update gives
\begin{equation}
    \Delta l_{\tau,\mathrm{raw}}^{(t)}
    \approx
    \alpha
    \left\langle
    \nabla_{\tau}^{(t)},
    \mathbf{u}_{\mathrm{raw}}^{(t)}
    \right\rangle .
    \label{eq:app_raw_target_change}
\end{equation}
Substituting Eq.~\eqref{eq:app_update_relation} into Eq.~\eqref{eq:app_target_change}, we have
\begin{equation}
\begin{aligned}
    \Delta l_{\tau}^{(t)}
    &\approx
    \alpha
    \left\langle
    \nabla_{\tau}^{(t)},
    \mathbf{u}_{\mathrm{raw}}^{(t)}
    -
    \beta c^{(t)}\nabla_r^{(t)}
    \right\rangle \\
    &=
    \Delta l_{\tau,\mathrm{raw}}^{(t)}
    -
    \alpha\beta c^{(t)}
    \left\langle
    \nabla_{\tau}^{(t)},
    \nabla_r^{(t)}
    \right\rangle .
\end{aligned}
\label{eq:app_target_decomposition}
\end{equation}
\begin{equation}
\begin{aligned}
    \Delta l_{\tau}^{(t)}
    -
    \Delta l_{\tau,\mathrm{raw}}^{(t)}
    \approx
    -
    \alpha\beta c^{(t)}
    \left\langle
    \nabla_{\tau}^{(t)},
    \nabla_r^{(t)}
    \right\rangle .
\end{aligned}
\label{eq:app_target_gain_condition}
\end{equation}
When the conditions in Eq.~\eqref{eq:app_sufficient_condition} hold, the right-hand side of Eq.~\eqref{eq:app_target_gain_condition} is positive. Hence $\Delta l_{\tau}^{(t)}>\Delta l_{\tau,\mathrm{raw}}^{(t)}$, so the corrected update improves the target-model misalignment score over the raw update. If $\left\langle\nabla_{\tau}^{(t)},\nabla_r^{(t)}\right\rangle=0$, the first-order effect is neutral. If $\left\langle\nabla_{\tau}^{(t)},\nabla_r^{(t)}\right\rangle>0$, the correction may reduce the target loss. This completes the proof.

We also provide a perturbation-level diagnostic using held-out white-box target models, where target gradients are used only for analysis. We compare the main adversarial image $\mathbf{x}_I+\boldsymbol{\delta}_m$ without debiasing with the final debiased adversarial image $\mathbf{x}_I+\boldsymbol{\delta}_I$ produced by DeBias-Attack. Table~\ref{tab:target_gradient_diagnostic} shows that DeBias-Attack gives higher target misalignment loss than the uncorrected main adversarial image. On TCL, the loss increases from $-0.7861$ to $-0.5272$, with a gain of $0.2589$ and a win rate of $99.1\%$. DeBias-Attack also improves the target loss by $0.0380$ on CLIP$_{\rm ViT}$ and $0.0344$ on CLIP$_{\rm CNN}$. The cosine between the debiasing component $\boldsymbol{\delta}_I-\boldsymbol{\delta}_m$ and the target gradient is small but positive, with mean values of $0.0082$, $0.0011$, and $0.0008$ on TCL, CLIP$_{\rm ViT}$, and CLIP$_{\rm CNN}$.

\begin{table}[!t]
\caption{Held-out target diagnostic comparing the main adversarial perturbation $\boldsymbol{\delta}_m$ without debiasing and the final debiased adversarial perturbation $\boldsymbol{\delta}_I$ of DeBias-Attack using ALBEF as the surrogate on Flickr30K. Higher loss and gain indicate stronger target misalignment.}
\label{tab:target_gradient_diagnostic}
\centering
\footnotesize
\renewcommand{\arraystretch}{1.08}
\setlength{\tabcolsep}{4.2pt}
\resizebox{\linewidth}{!}{%
\begin{tabular}{lccccc}
\toprule
Target &
$\cos(\nabla_{\tau}, \boldsymbol{\delta}_I-\boldsymbol{\delta}_m)$ & $\boldsymbol{\delta}_m$ & $\boldsymbol{\delta}_I$ &
Gain &
Win rate \\
\midrule
TCL & 0.0082 & -0.7861 & -0.5272 & 0.2589 & 99.1 \\
CLIP$_{\rm ViT}$ & 0.0011 & -1.3857 & -1.3477 & 0.0380 & 75.9 \\
CLIP$_{\rm CNN}$ & 0.0008 & -2.2037 & -2.1693 & 0.0344 & 79.1 \\
\bottomrule
\end{tabular}}
\end{table}

\begin{figure*}[!t]
    \centering
    \includegraphics[width=0.80\linewidth]{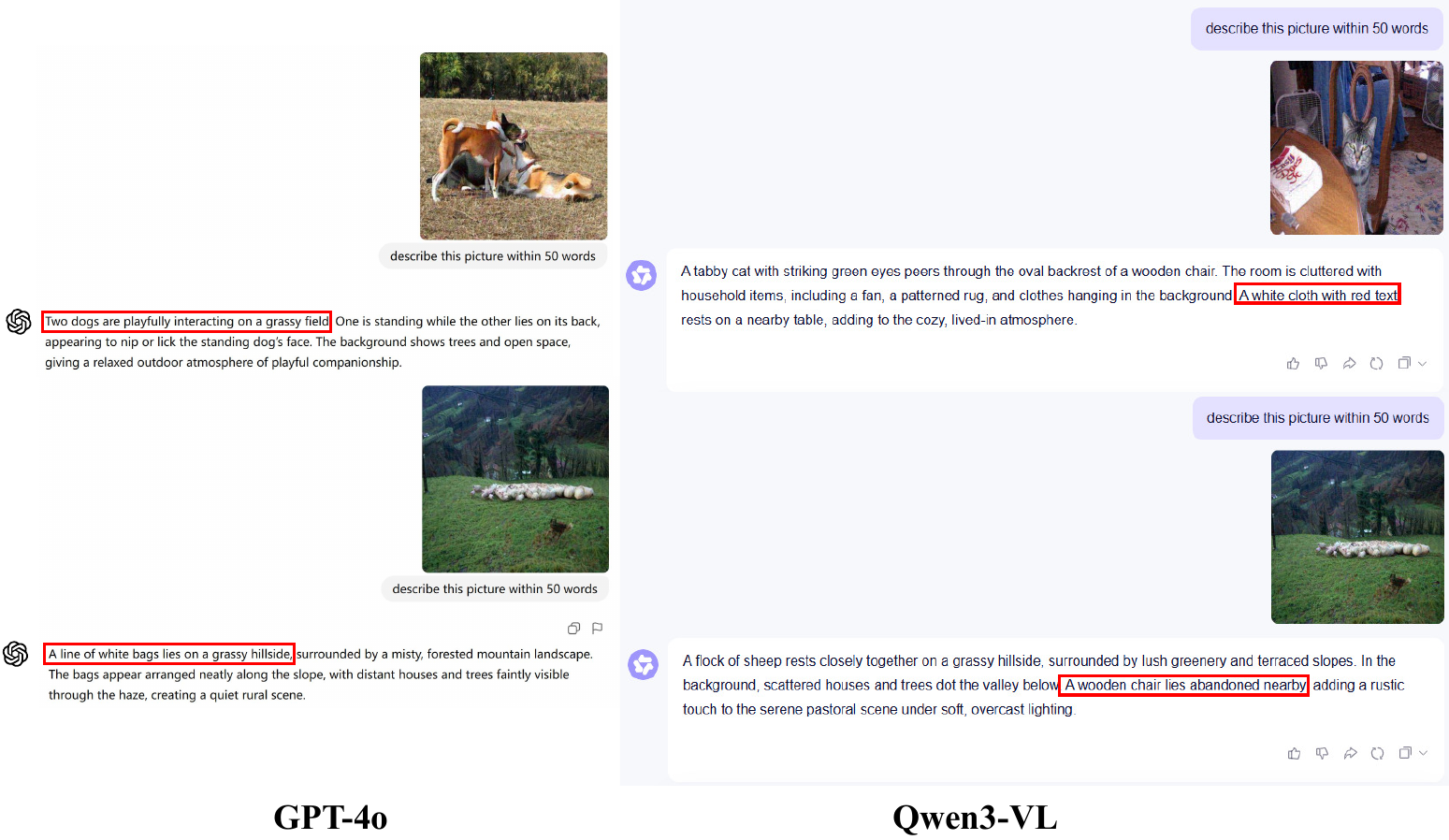}
    \vspace{-3pt}
    \caption{Qualitative MLLM transfer examples. The examples compare responses before and after attack; misleading or incomplete adversarial responses indicate stronger transfer.}
    \label{fig:performance_mllm}
\end{figure*}

The reference adversarial gradient is estimated from weak-semantic images and evaluated under the original-image context, so it can expose surrogate-preferred directions. By updating the final debiased perturbation $\boldsymbol{\delta}_I$ with the corrected direction, DeBias-Attack reduces dependence on the uncorrected main perturbation $\boldsymbol{\delta}_m$ while preserving adversarial optimization. The perturbation-level diagnostic shows higher target-side misalignment than the uncorrected main perturbation on held-out target models.

\section{Qualitative Adversarial Examples}
\label{app:qualitative}

Figure~\ref{fig:performance_mllm} shows qualitative examples of MLLM attack transfer on GPT-4o and Qwen3-VL. As in the quantitative setting, these adversarial examples are generated on ALBEF. After attack, reasonable descriptions are replaced by misleading, incomplete, or hallucinatory content. Some responses still preserve partial semantics, but the examples match the empirical transfer results.

\section{Attention Visualization and Modality Ablation}
\label{app:attention_ablation}

\begin{figure*}[!t]
    \centering
    \includegraphics[width=0.9\linewidth]{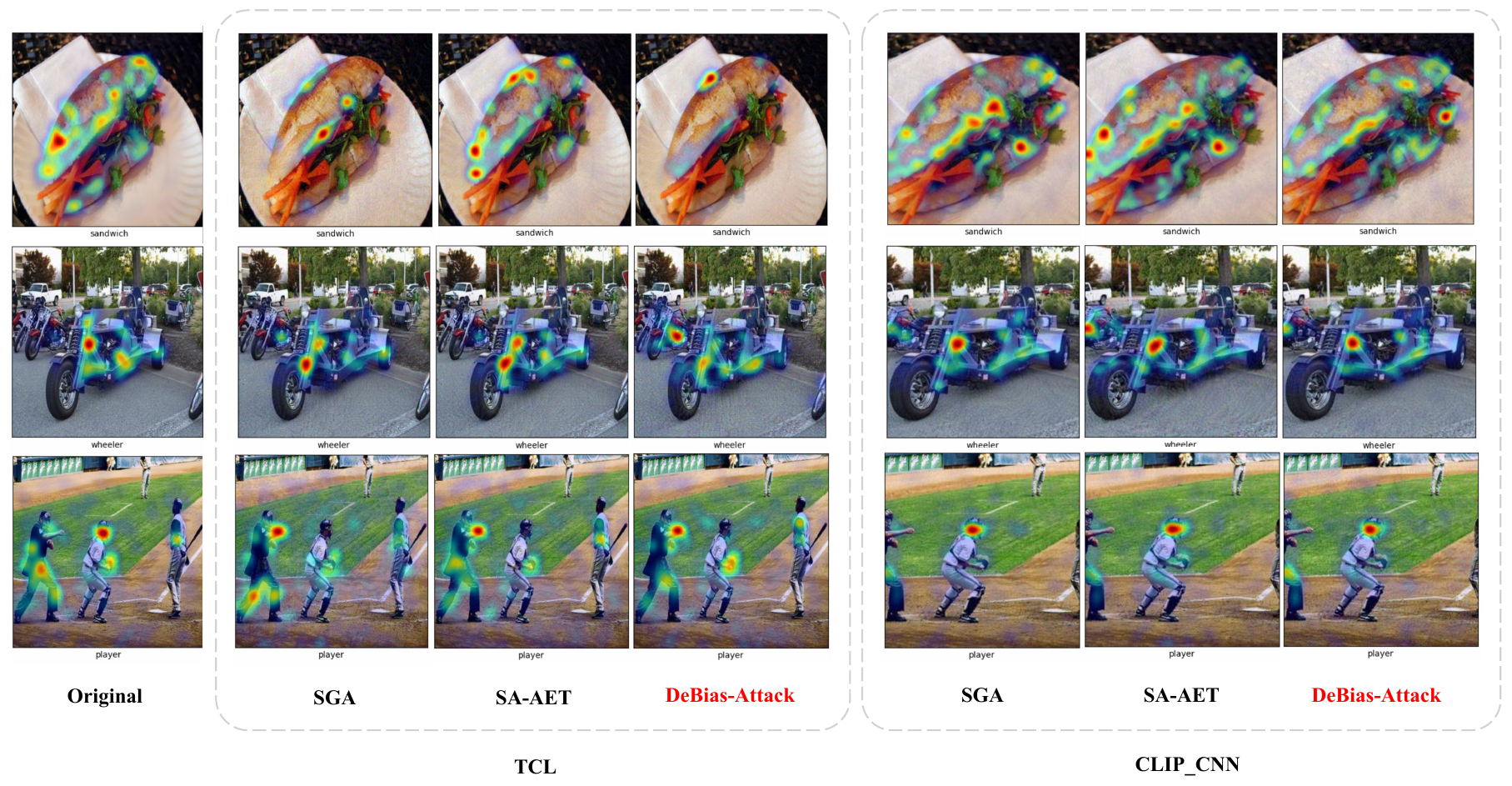}
    \vspace{-3pt}
    \caption{Vision-grounding attention maps under attack. Brighter regions indicate stronger attention, and dispersed or shifted attention suggests stronger disruption of visual grounding.}
    \label{fig:vision_cam}
\end{figure*}

Figure~\ref{fig:vision_cam} visualizes attention maps under different attack methods. A successful adversarial example should disrupt the model's focus on correct grounding regions, producing dispersed or misaligned attention. All adversarial examples are crafted using ALBEF as the surrogate model and evaluated on two targets: TCL, which shares a similar architecture with ALBEF, and CLIP$_\text{CNN}$, which differs in design. Compared with SA-AET and SGA, DeBias-Attack induces broader and less concentrated attention, indicating stronger interference with grounding behavior. On TCL, SA-AET retains partial focus near the main object, whereas DeBias-Attack diffuses attention toward background regions. On CLIP$_\text{CNN}$, SGA maintains some activation on target-relevant areas, but DeBias-Attack scatters attention across peripheral or irrelevant regions, such as surrounding textures or secondary objects.

DeBias-Attack alters prediction outputs and degrades the model's internal localization behavior more than prior approaches. The attention dispersion matches the design: the debiased perturbation is updated with a corrected direction that reduces surrogate-specific bias while preserving cross-modal attack strength.

\begin{figure*}[!t]
    \centering
    \includegraphics[width=0.9\linewidth]{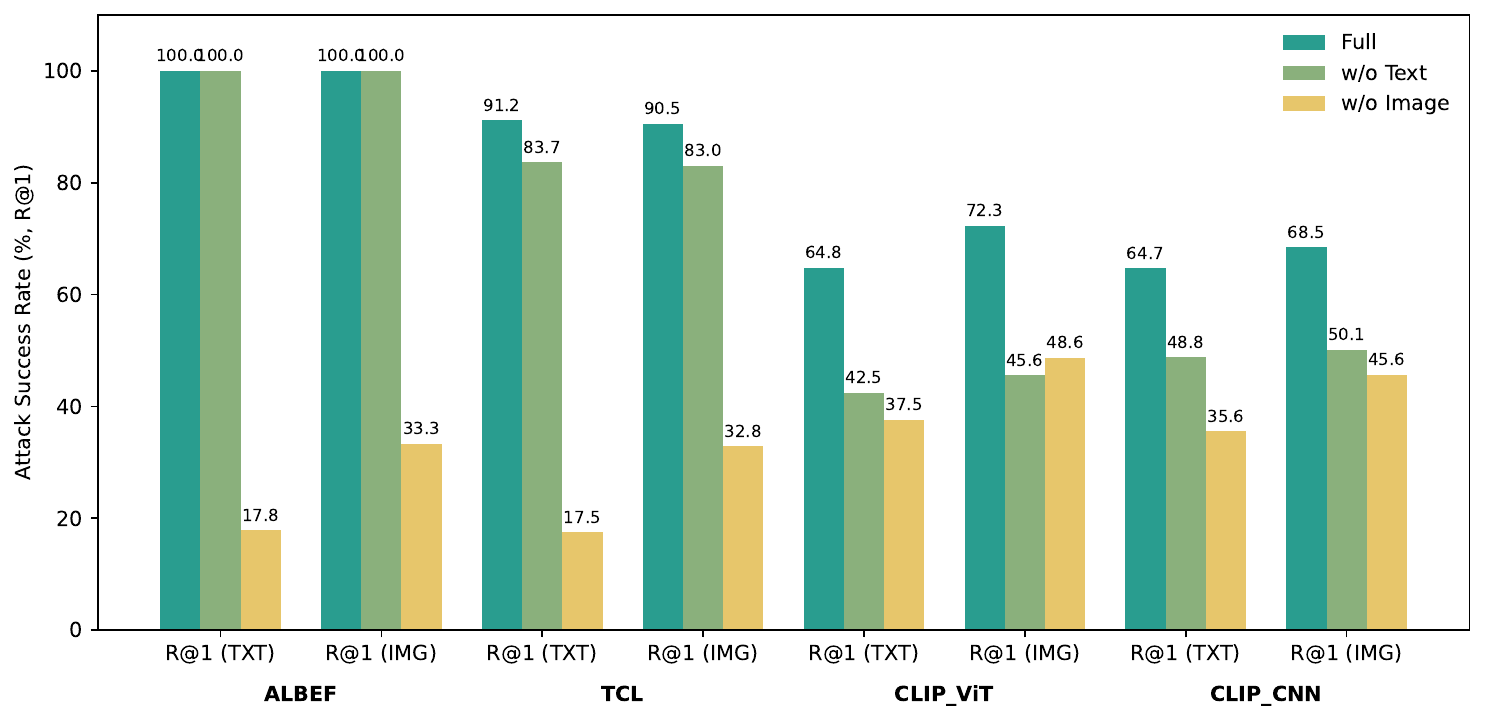}
    \vspace{-3pt}
    \caption{Modality-level ablation on retrieval tasks. Higher ASR indicates a stronger attack, and comparing variants shows the contribution of each optimized modality.}
    \label{fig:ablation_attacker}
\end{figure*}

We conduct a modality-level ablation study on Flickr30K using ALBEF as the surrogate model. Figure~\ref{fig:ablation_attacker} reports ASR for text retrieval and image retrieval across target models under the full, w/o text, and w/o image settings. The full configuration achieves the highest ASR across all models. For related models such as ALBEF and TCL, removing image-side perturbations causes a larger performance drop, indicating that visual perturbations dominate transfer when the models share similar alignment behavior. For heterogeneous models such as CLIP$_{\rm CNN}$ and CLIP$_{\rm ViT}$, the gap between the two variants narrows, suggesting that both modalities contribute to transferable cross-modal misalignment. Image-side correction and text-side substitution are complementary: the image update reduces surrogate-specific bias, while the text update strengthens semantic misalignment with the updated adversarial image.

\section{Defense Evaluation}
\label{app:defense}

\begin{table}[!t]
\caption{Comparison of different defense methods on MSCOCO. Each cell reports defended ASR and the change relative to the undefended setting; lower ASR ($\downarrow$) and larger positive reductions indicate stronger mitigation.}
\centering
\small
\renewcommand\arraystretch{1.1}
\setlength{\tabcolsep}{6pt}
\resizebox{0.85\linewidth}{!}{
\begin{tabular}{c|c|cc}
\toprule
\multirow{2}{*}{Models} & \multirow{2}{*}{Defense Method} & \multicolumn{2}{c}{MSCOCO} \\
\cmidrule(lr){3-4}
 & & TR$\downarrow$ & IR$\downarrow$ \\
\midrule

\multirow{2}{*}{ALBEF} & RT & 99.68/\textcolor{red}{0.32} & 99.60/\textcolor{red}{0.40} \\
& LT & 99.92/\textcolor{red}{0.08} & 99.84/\textcolor{red}{0.16} \\
\midrule
\multirow{2}{*}{TCL} & RT & 88.91/\textcolor{red}{1.29} & 89.85/\textcolor{red}{0.30} \\
& LT & 93.43/\textcolor{red}{-3.24} & 93.68/\textcolor{red}{-3.53} \\
\midrule
\multirow{2}{*}{CLIP$_{\text{ViT}}$} & RT & 69.74/\textcolor{red}{4.07} & 75.75/\textcolor{red}{1.54} \\
& LT & 70.77/\textcolor{red}{3.04} & 76.73/\textcolor{red}{0.57} \\
\midrule
\multirow{2}{*}{CLIP$_{\text{CNN}}$} & RT & 72.12/\textcolor{red}{0.07} & 77.01/\textcolor{red}{0.37} \\
& LT & 72.35/\textcolor{red}{-0.17} & 77.34/\textcolor{red}{0.05} \\
\bottomrule
\end{tabular}}
\label{tab:defense}
\end{table}

We evaluate DeBias-Attack under two test-time defenses. Random transform (RT) applies stochastic image transformations at inference and aggregates similarities. LanguageTool (LT)~\cite{morris2020reevaluating} performs grammar-based text normalization to sanitize perturbed tokens. We report ASR for TR and IR. In Table~\ref{tab:defense}, each cell shows the defended ASR and its change relative to the undefended model. A larger positive change indicates stronger mitigation, whereas a negative change means that the defense fails to reduce the attack and can even increase ASR by a small amount. DeBias-Attack survives both defenses across models, so these off-the-shelf defenses provide limited protection against transferable cross-modal semantic misalignment.

\end{document}